\documentclass[11pt]{article}

\usepackage[preprint]{acl}
\usepackage{times}
\usepackage{latexsym}

\usepackage[T1]{fontenc}
\usepackage[utf8]{inputenc}

\usepackage{microtype}

\usepackage{inconsolata}

\usepackage{amsfonts}
\usepackage{amsmath}
\usepackage{amssymb}
\usepackage{nicefrac}
\usepackage{booktabs}
\usepackage{tabularx}
\usepackage{multirow}
\usepackage[table]{xcolor}
\usepackage{graphicx}
\usepackage{tikz}
\usetikzlibrary{arrows.meta,positioning,shapes.geometric,calc,fit,backgrounds}
\usepackage{algorithm}
\usepackage{algpseudocode}
\usepackage{xspace}
\usepackage{pifont}
\usepackage{tcolorbox}
\tcbuselibrary{breakable, skins}

\newcommand{\method}{Visual Para-Thinker++\xspace}
\newcommand{\xmark}{\ding{55}}
\newcommand{\cmark}{\ding{51}}

\newtcolorbox{sftbox}[1][]{
    breakable, enhanced jigsaw,
    colback=white, colframe=black!75, coltitle=white, fonttitle=\bfseries,
    title=Data Example (Block-based allocation), arc=3pt, boxrule=0.8pt,
    toptitle=2pt, bottomtitle=2pt, left=8pt, right=8pt, top=6pt, bottom=6pt
}
\newtcolorbox{sftboxone}[1][]{
    breakable, enhanced jigsaw,
    colback=white, colframe=black!75, coltitle=white, fonttitle=\bfseries,
    title=Data Example (Scan-order allocation), arc=3pt, boxrule=0.8pt,
    toptitle=2pt, bottomtitle=2pt, left=8pt, right=8pt, top=6pt, bottom=6pt
}

\title{\method: A Single-Policy Multi-Agent Framework for Visual Reasoning}

\author{
  Haoran Xu$^{1}$\thanks{Equal contribution.} \quad Hongyu Wang$^{2}$\footnotemark[1] \quad Yifei Gao$^{3}$\footnotemark[1] \quad Jiaze Li$^{1}$\thanks{Project leader.} \quad Zizhao Tong$^{4}$\\  \textbf{Xiaofeng Zhang}$^{5}$\thanks{Corresponding author.} \quad \textbf{Xiaosong Yuan}$^{6}$\footnotemark[3] \\
  $^{1}$Zhejiang University 
  $^{2}$Hunan University
  $^{3}$Tianjin University\\
  $^{4}$University of Chinese Academy of Sciences\\
  $^{5}$Shanghai Jiao Tong University
  $^{6}$Jilin University\\
  \small{
    \textbf{Correspondence:} \href{mailto:xhr964691257@163.com}{xhr964691257@163.com}
  }
}

\begin{document}
\maketitle

% =====================================================================
\begin{abstract}
Visual reasoning requires integrating evidence distributed across regions, attributes, and relations, making single-chain reasoning prone to early perceptual commitment and hallucination. We propose \method, a single-policy multi-agent framework in which one shared MLLM policy is instantiated as role-conditioned Main, Worker, and Summary Agents. The Main Agent decomposes the task with fixed allocation patterns; Worker Agents reason in parallel under context isolation; and the Summary Agent reconciles full Worker reasoning traces rather than majority-voting on final labels. The shared policy is trained by Multi-Agent Capability Injection and Role-Decoupled Multi-Agent Optimization, which assign role-specific rewards and advantages to corresponding token segments to reduce gradient conflict among collaborative roles. A native inference engine enables efficient multi-agent rollout through shared visual prefix and KV cache reuse. Across V\textsuperscript{*}, CountBench, Pixmo, MMVP, the RefCOCO family, and HallusionBench, \method consistently outperforms single-trajectory and inference-time parallel baselines, with especially strong gains on hallucination-sensitive visual reasoning.
\end{abstract}

% =====================================================================
\section{Introduction}
\label{sec:intro}

Test-time scaling has become the dominant lever for improving the reasoning ability of large language models (LLMs)~\citep{wei2022chain,openai2024o1}: lengthen a single chain of thought, encourage self-reflection, and let one trajectory revise its own intermediate conclusions. Reinforcement Learning with Verifiable Rewards (RLVR) further amplifies this single-agent recipe by optimizing one policy against an outcome signal. The recipe has been effective in text-dominant domains such as mathematics and code, but its returns are markedly weaker on visual reasoning, where relevant evidence is distributed across regions, attributes, and relations rather than concentrated in a single linguistic argument. A single trajectory typically commits to one perceptual interpretation early; once an interpretation is selected, lengthening the chain tends to deepen it rather than diversify it, and an early perceptual error tends to be justified rather than corrected.

This observation motivates a different scaling axis: rather than making one agent's chain longer, organize visual reasoning as multi-role collaboration. Different reasoning roles, including question decomposition, local evidence gathering, global count verification, and intermediate finding reconciliation, are naturally assigned to different agents. Based on above, we propose a multi-role collaboration framework, termed \method. Rather than using multiple independent models—which is computationally expensive—we introduce a single-policy multi-agent system. By conditioning a shared MLLM on specific role tokens, our approach enables subagents to collaborate.

Specifically, \method multi-agent framework consists of three types of roles:
\begin{itemize}
\item \textbf{Main Agent} that, given the image and the question, the role decomposes the visual task into sub-tasks based on two distinct allocation patterns, assigning each to worker agents.

\item \textbf{Worker Agents}, upon receiving sub-tasks from the main agent, worker agents execute their respective assignments in a decoupled and isolated manner.

\item \textbf{Summary Agent} that aggregates the outputs from various sub-agents, integrating their respective reasoning results to formulate a coherent final response.
\end{itemize}

To realize this multi-agent collaboration within a single policy, we propose Role-Decoupled Multi-Agent Optimization. First, we employ Multi-Agent Capability Injection, a role-aware SFT objective that enables a shared policy to instantiate distinct Main, Worker, and Summary roles. We utilize a context-isolation mask to ensure that Worker Agents operate independently, while the Summary Agent maintains full visibility of all preceding traces. Subsequently, the policy is refined via Role-Decoupled Optimization, where rewards are decomposed into role-specific signals: Worker rewards evaluate local evidence and sub-task accuracy, while the Summary reward assesses the final output. By mapping these signals to role-specific token segments, we mitigate gradient conflicts arising from the divergent objectives of the agents—specifically, the tension between local evidence gathering and global synthesis—while maintaining a unified set of model weights.

Empirically, \method consistently outperforms both single-trajectory (e.g., greedy decoding, long CoT) and parallel inference baselines (e.g., self-consistency, multi-agent debate) across six visual benchmarks. Finally, to enhance inference efficiency, we optimize the system's execution performance by leveraging KV cache reuse across the multi-agent reasoning process~\cite{ye2025kvcommonlinecrosscontextkvcache}.

\paragraph{Contributions.} Our contributions are summarised in three points:
\begin{itemize}
\item We propose \method, a single-policy multi-agent visual reasoning framework where one shared MLLM policy is instantiated as Main Agent, Worker Agents, and Summary Agent through role conditioning.

\item We introduce a unified two-stage optimization for \method: Multi-Agent Capability Injection and Role-Decoupled Multi-Agent Optimization to reduce the gradient conflicts among different roles.

\item Across V\textsuperscript{*}, CountBench, Pixmo, MMVP, the RefCOCO family, and HallusionBench, \method improves visual reasoning more effectively than extending a single chain of thought. Practical evaluation is made efficient by a native vLLM-based multi-agent rollout engine that reuses KV cache across roles.
\end{itemize}

% =====================================================================
\section{Related Work}
\label{sec:related}

\paragraph{Reasoning MLLMs.}
Since GPT-4V, MLLMs have been asked to produce visible chains-of-thought for visual question answering~\citep{zhang2023multimodal,liu2024llava}. Recent work either post-trains MLLMs with reasoning data (LLaVA-o1~\citep{xu2024llavao1}, Visual-CoT~\citep{shao2024visualcot}) or activates reasoning through test-time prompting~\citep{mitra2024compositional,liu2025efficientreasoningsuppressionselfaffirmation,yuan2026differential}. These are fundamentally single-agent, single-trajectory systems; they do not explicitly assign reasoning roles to different agents.

\paragraph{Parallel and self-consistent reasoning.}
Self-consistency~\citep{wang2023selfconsistency} samples $K$ independent chains and majority-votes on the final answer. Tree/Graph-of-Thought~\citep{yao2023tree} generalize to search trees, and Para-Thinker~\citep{pan2025parathinker} explores shared prefixes with divergent suffixes in a text-only setting. These methods are inference-time heuristics drawn from the same single role distribution: every chain inherits the same single-agent bias, and the system aggregates only at the final-answer level. Our framework instead learns role-conditioned agents into a single shared policy and aggregates at the trace level via a Summary Agent.

\paragraph{Multi-agent LLMs.}
Multi-agent debate~\citep{du2023improving} and role-play systems~\citep{chen2023agentverse} instantiate multiple distinct models or prompts to encourage role separation. The dominant cost is $K$-fold inference on heterogeneous models, and gradients rarely flow through the team. Our single-policy multi-agent design collapses these $K$ models into one shared set of weights, makes role identity a function of role tokens and visible context rather than of separate models, and admits end-to-end optimization within a unified RL stage.

\begin{figure*}[t]
\centering
\includegraphics[width=0.95\textwidth]{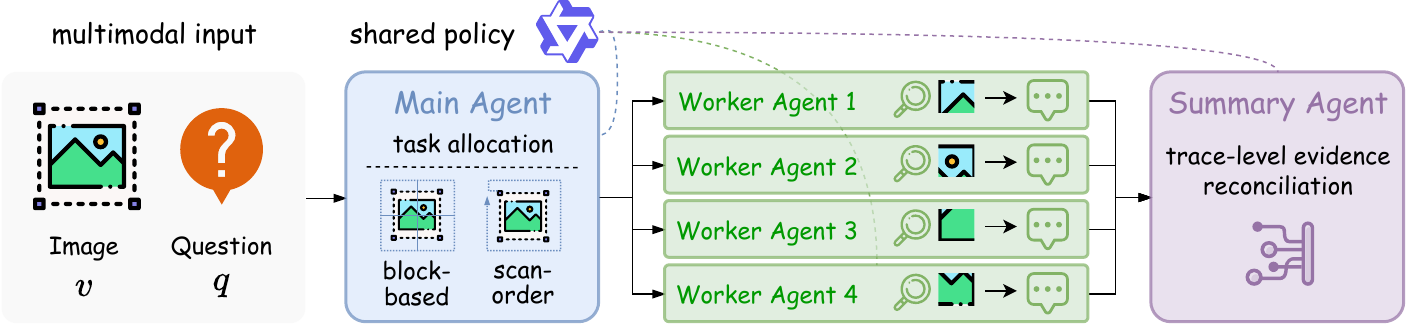}
\caption{\method framework. A single shared policy $\pi_\theta$ is instantiated as three role-conditioned agents under a fixed protocol. Given a multimodal input $(v,q)$, the \textbf{Main Agent} emits one of two fixed task-allocation patterns (block-based or scan-order); four \textbf{Worker Agents}, role-conditioned instances of the same policy, then explore the dispatched visual sub-problems along parallel reasoning paths under mutually isolated contexts; a \textbf{Summary Agent} performs trace-level evidence reconciliation to produce the final answer. All three agents share the same policy weights (see the shared-policy icon on top); only the role embedding and the visible context differ.}
\label{fig:framework}
\end{figure*}
\begin{figure}[t]
\centering
\includegraphics[width=0.8\linewidth]{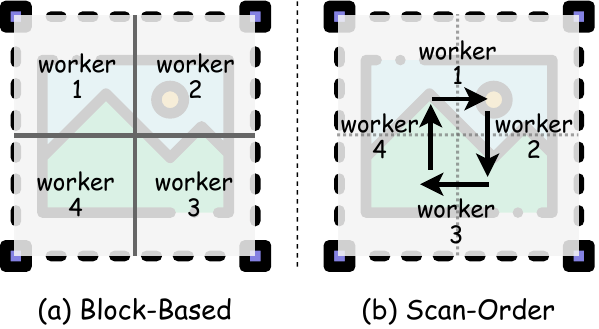}
\caption{The two fixed task-allocation patterns used by the Main Agent under the four-Worker protocol. (a) \textbf{Block-based}: the image is partitioned into four disjoint quadrants, each dispatched to one Worker Agent for local evidence collection. (b) \textbf{Scan-order}: each Worker Agent retains a global view but traverses the image along a distinct scan order.}
\label{fig:scan}
\end{figure}

% =====================================================================
\section{Method}
\label{sec:method}

\subsection{Overview}
\label{sec:overview}

\method instantiates one shared MLLM policy as three role-conditioned agents under a fixed four-path collaboration protocol (Fig.~\ref{fig:framework}). The Main Agent receives the image and the question and decomposes the visual task into complementary sub-problems through a small set of fixed task-allocation patterns. A fixed set of Worker Agents, role-conditioned instances of the same policy, then explore these sub-problems along parallel reasoning paths under context isolation. A Summary Agent reads the complete Worker reasoning traces and produces the final answer through trace-level evidence reconciliation.

We deliberately frame the Main Agent as a fixed-pattern dispatcher rather than as an open-ended planner, and the Summary Agent as a learned reconciler rather than as a majority voter. There are no independently trained policies: all three agents share the same set of weights $\theta$, and role identity is induced solely by role tokens and visible context. \method is therefore a single-policy multi-agent system: a unified optimization framework rather than a collection of agents glued together by prompts.

% -------- Figure 2: framework --------
\begin{figure*}[t]
\centering
\includegraphics[width=0.95\textwidth]{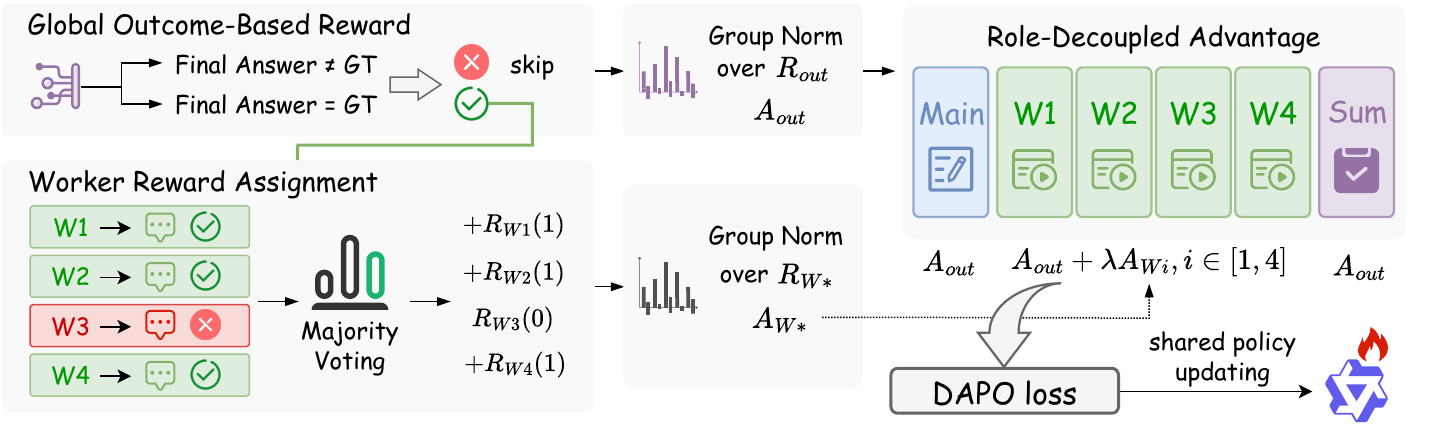}
\caption{Training pipeline of Role-Decoupled Multi-Agent Optimization. \textbf{Top branch (Global Outcome-Based Reward).} A trajectory contributes the outcome reward $R_\text{out}\!\in\!\{0,1\}$ depending on whether the Summary Agent's final answer matches ground truth; group normalisation over the on-policy group then yields the outcome advantage $A_\text{out}$. \textbf{Bottom branch (Worker Reward Assignment).} Each Worker Agent's intermediate sub-answer is graded by a lightweight cross-Worker majority-vote heuristic (e.g., W3 disagrees with the majority and is penalised, $R_{W3}{=}0$); group normalisation over all Worker rewards yields the per-Worker advantage stream, written compactly as $A_{W*}$ in the figure and instantiated as $A_w^{(i)}$ for Worker $i$. \textbf{Composition.} The two advantage streams are composed token-wise (Eq.~\ref{eq:adv}): Main and Summary segments receive $A_t{=}A_\text{out}$, while Worker $i$'s segment receives $A_t{=}A_\text{out}+\lambda A_w^{(i)}$. The composed advantages are plugged into a clipped DAPO loss to update the shared policy.}
\label{fig:rdmao-pipeline}
\end{figure*}

\subsection{Single-Policy Multi-Agent Formulation}
\label{sec:formulation}

We model \method as a single autoregressive policy $\pi_\theta$ unrolled under a role-aware state machine. Given image $v$ and question $q$, a multi-agent trajectory $\tau$ is decomposed into $K{+}2$ role segments:
\begin{equation}
  \tau = \underbrace{m}_{\text{Main}}\!\Big\|\!\underbrace{s_1}_{\text{Worker}_1}\!\Big\|\cdots\Big\|\!\underbrace{s_K}_{\text{Worker}_K}\!\Big\|\!\underbrace{u}_{\text{Summary}},
  \label{eq:traj}
\end{equation}
where the Main segment $m$ is conditioned on $(v, q, r_\text{main})$, each Worker segment $s_i$ on $(v, q, m, r_{\text{worker}_i})$, and the Summary segment $u$ on $(v, q, m, s_{1:K}, r_\text{sum})$. Worker Agents never see each other's tokens during their own reasoning; the Summary Agent retains full visibility. Role tokens $r_{(\cdot)}$ are learned embeddings in one shared vocabulary; all three agents share the transformer weights $\theta$. The protocol order Main\,$\to$\,Worker\,$\to$\,Summary, the four-path layout, and the visibility relations are fixed at training and inference.

\subsection{Main-Agent Task Allocation with Fixed Patterns}
\label{sec:mainagent}

The Main Agent converts $(v,q)$ into $K$ visual sub-problems for the Worker Agents through one of two fixed allocation patterns (Fig.~\ref{fig:scan}): \textbf{Block-based allocation} partitions the image into $K$ disjoint regions and dispatches each Worker Agent to collect local evidence, fitting grounding, fine-grained perception, and attribute verification; \textbf{Scan-order allocation} assigns each Worker Agent a distinct global scanning trajectory while retaining a full receptive field, fitting counting and global verification where errors arise from missing or double-counting under a single direction. By default the Main Agent emits Block-based allocation and switches to Scan-order when the task is dominated by counting or global verification; the chosen pattern is part of segment $m$ and is visible to all Workers and to the Summary Agent.

\subsection{Multi-Agent Capability Injection}
\label{sec:sft}

The first training stage teaches the shared policy $\pi_\theta$ to instantiate Main, Worker, and Summary roles via SFT. Given a sample $(v, q, m^\star, s_{1:K}^\star, u^\star)$ synthesised by a stronger MLLM teacher (Appendix~\ref{app:hparam}), we minimise
\begin{equation}
\begin{aligned}
\mathcal{L}_\text{MAC}(\theta)
&= - \!\!\sum_{t\in m}\!\! \log \pi_\theta(m_t \mid c_m, m_{<t}) \\
&\quad - \sum_{i=1}^K \sum_{t\in s_i} \log \pi_\theta(s_{i,t} \mid c_i, s_{i,<t}) \\
&\quad - \sum_{t\in u} \log \pi_\theta(u_t \mid c_u, u_{<t}),
\end{aligned}
\label{eq:sft}
\end{equation}
with $c_m{=}(v,q,r_\text{main})$, $c_i{=}(v,q,m,r_{\text{worker}_i})$, and $c_u{=}(v,q,m,s_{1:K},r_\text{sum})$. The role tokens $r_{(\cdot)}$ are the only inputs distinguishing the three agents under shared weights, and the Worker index $i$ in $r_{\text{worker}_i}$ forces $\pi_\theta(\cdot\mid\cdot,r_{\text{worker}_i})$ to specialise across $i$ rather than collapse. We additionally apply a context-isolation mask on the packed trajectory so that each Worker Agent attends only to $(v,q,m)$ plus its own suffix while the Summary Agent retains full visibility of all Worker traces; without this mask the parallel paths copy each other rather than explore complementary evidence, and the Summary Agent's full visibility is what later enables trace-level reconciliation in \S\ref{sec:rdmao}.

\subsection{Role-Decoupled Multi-Agent Optimization}
\label{sec:rdmao}

The second stage refines the shared policy with group-relative RL over the multi-agent trajectory of Eq.~\ref{eq:traj}. Worker Agents and the Summary Agent pursue partially different objectives, so a single trajectory-level reward broadcast to every token couples their gradients: a wrong final answer poisons every Worker, and a noisy Worker reward leaks into the Summary segment. We introduce role-specific rewards together with role-specific advantages applied only to the corresponding token segments; the overall pipeline is shown in Fig.~\ref{fig:rdmao-pipeline}.

% -------- Figure: RDMAO training pipeline --------
\begin{figure*}[t]
\centering
\includegraphics[width=0.98\textwidth]{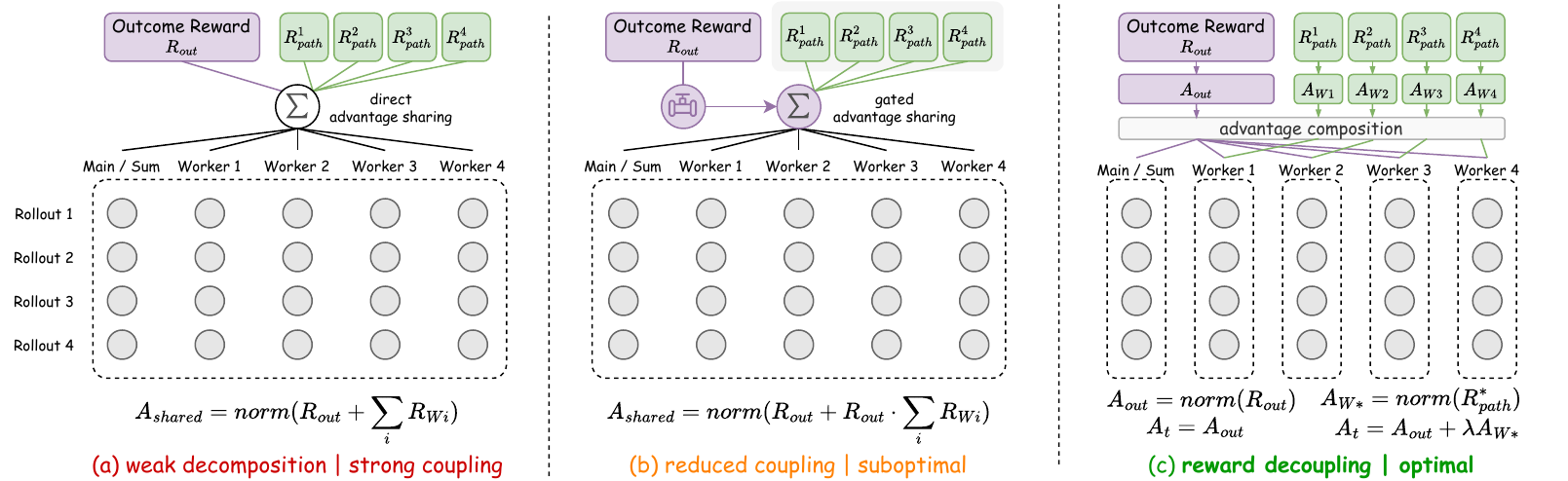}
\caption{Three representative ways to compose a Worker reward with the outcome reward across a multi-agent rollout. The figure denotes the per-Worker reward as $R_{Wi}$ (also written as $R_\text{path}^i$) and its normalised advantage as $A_{W*}$; in the main text we equivalently write these as $R_w^{(i)}$ and $A_w^{(i)}$. \textbf{(a) Naive Sum} (\emph{weak decomposition, strong coupling}): all Worker rewards are summed with $R_\text{out}$ into a single $A_\text{shared}=\mathrm{norm}(R_\text{out}+\sum_i R_{Wi})$ that is broadcast to every token segment, coupling Worker and outcome gradients. \textbf{(b) Conditional Sum} (\emph{reduced coupling, suboptimal}): $R_\text{out}$ gates the Worker rewards via $A_\text{shared}=\mathrm{norm}(R_\text{out}+R_\text{out}\!\cdot\!\sum_i R_{Wi})$, but the advantage is still shared across roles. \textbf{(c) Role-decoupled Advantages (ours)} (\emph{reward decoupling, optimal}): $R_\text{out}$ and each $R_{Wi}$ are normalised \emph{within their own on-policy groups} into $A_\text{out}$ and $A_{W*}$, then composed token-wise via Eq.~\ref{eq:adv}---Main and Summary tokens receive $A_t=A_\text{out}$, while Worker $i$'s tokens receive $A_t=A_\text{out}+\lambda A_{W*}$. }
\label{fig:reward}
\end{figure*}

\paragraph{Reward signals.}
Each Worker Agent $i$ receives a worker-level reward $R_\text{worker}^{(i)}\in[0,1]$ that grades its local reasoning trace, instantiated by a lightweight cross-Worker majority-vote heuristic and used only as a training signal. The team-level outcome reward $R_\text{out}(\tau)\in\{0,1\}$ grades the Summary Agent's final answer and supervises the Main and Summary segments.

\paragraph{Role-decoupled advantages.}
We group-normalise each reward source over the on-policy DAPO group separately:
\begin{equation}
\begin{aligned}
A_\text{out} &= \frac{R_\text{out} - \mu_\text{out}}{\sigma_\text{out}}, \\
A_\text{worker}^{(i)} &= \frac{R_\text{worker}^{(i)} - \mu_\text{worker}}{\sigma_\text{worker}},
\end{aligned}
\label{eq:rew-norm}
\end{equation}
and compose them token-wise:
\begin{equation}
A_t =
\begin{cases}
A_\text{out}, & t \in m \cup u, \\
A_\text{out} + \lambda\, A_\text{worker}^{(i)}, & t \in s_i,
\end{cases}
\label{eq:adv}
\end{equation}
with Worker weight $\lambda{=}0.5$ (Appendix~\ref{app:abl}). The outcome advantage flows to every role as the team-level signal, while the Worker advantage updates only its own segment, equivalent to a segment-wise mask on the gradient: a mistake in Worker $j$ changes the advantage only of tokens in $s_j$. The advantages are plugged into a clipped DAPO~\citep{yu2025dapo} surrogate loss.

\paragraph{How to compose $R_w$ and $R_o$.}
A natural design question is \emph{at which level} the Worker reward $R_w$ should be combined with the outcome reward $R_o$. We identify three representative schemes (Fig.~\ref{fig:reward}). \textbf{(a) Naive Sum} forms $R_o + \sum_i R_w^{(i)}$ and broadcasts a single normalised advantage to every token. \textbf{(b) Conditional Sum} gates the Worker rewards by the outcome via $R_o + R_o\!\cdot\!\sum_i R_w^{(i)}$ but still uses a shared advantage. \textbf{(c) Role-decoupled Advantages (ours)}, in the spirit of per-reward group normalisation for multi-reward RL~\citep{nvidia2026gdpo}, normalises $R_o$ and each $R_w^{(i)}$ within their own groups (Eq.~\ref{eq:rew-norm}) and composes them at the advantage level (Eq.~\ref{eq:adv}). Schemes (a) and (b) couple the two signals through a shared advantage; only (c) routes each reward source to the role tokens it is meant to supervise. Table~\ref{tab:ablation-composition} shows that (c) outperforms both summation variants under the same $R_w$, and Table~\ref{tab:ablation-reward} further shows that $R_w$ and the role-decoupled advantage are complementary, with the largest gains obtained by combining them.

% -------- Figure 4: reward composition paradigms --------

\subsection{Native Multi-Agent Inference Engine}
\label{sec:engine}

\begin{figure}[t]
\centering
\includegraphics[width=0.9 \linewidth]{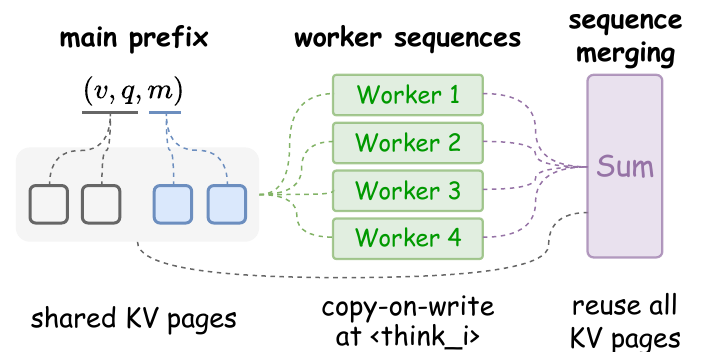}
\caption{Native multi-agent inference engine.}
\label{fig:engine}
\end{figure}

Efficient rollout is necessary for making single-policy multi-agent reasoning practical. A naive Main/Worker/Summary rollout requires $1{+}K{+}1$ decoding calls with redundant visual prefills. Our vLLM-based engine instead pays the image-question-Main prefill once, stores the shared KV pages in a parent \texttt{SequenceGroup}, forks $K$ Worker sequences at role-trigger tokens via PagedAttention Copy-on-Write, decodes Workers in parallel, and merges their KV pages before the Summary segment (Fig.~\ref{fig:engine}). Thus, Worker Agents share the visual prefix without copying memory at fork time, and the Summary Agent reuses both parent and Worker KV pages rather than refilling the complete multi-agent context.

% -------- Figure 3: engine --------

\begin{table*}[t]
  \caption{Comprehensive evaluation of models on vision-centric tasks, including counting, fine-grained perception, and hallucination tasks. The first \textbf{Avg.} column averages the three counting columns (Pixmo val, Pixmo test, CountBench); the second \textbf{Avg.} column averages V\textsuperscript{*}, MMVP, and HallusionBench. \textbf{Bold} marks our \method results in the last two rows; ``\,--\,'' denotes results not reported by the source. ``Majority voting@4'' generates four trajectories per query and aggregates by majority vote. \method extends Visual Para-Thinker with Role-Decoupled Multi-Agent Optimization, and improves over every Peer-Model baseline (the open-source 3B/7B group) on every benchmark and at both scales.}
  \label{tab:main}
  \centering
  \footnotesize
  \setlength{\tabcolsep}{3pt}
  \resizebox{\textwidth}{!}{%
  \begin{tabular}{@{}l|c|cc|c|c|c|c|c|c@{}}
    \toprule
    \multirow{3}{*}{\textbf{Model}} & \multirow{3}{*}{\textbf{Size}} & \multicolumn{3}{c|}{\textbf{Counting}} & \multirow{3}{*}{\textbf{Avg.}} & \textbf{Visual Search} & \textbf{Fine-grained} & \textbf{Hallucination} & \multirow{3}{*}{\textbf{Avg.}} \\ \cmidrule(lr){3-5} \cmidrule(lr){7-7} \cmidrule(lr){8-8} \cmidrule(lr){9-9}
    & & \multicolumn{2}{c|}{Pixmo} & CountBench & & \multirow{2}{*}{\textbf{V*}} & \multirow{2}{*}{MMVP} & \multirow{2}{*}{HallusionBench} & \\ \cmidrule(lr){3-4} \cmidrule(lr){5-5}
    & & val & test & test & & & & & \\ \midrule
    \rowcolor[HTML]{F2F2F2} \textit{Frontier Models} & & & & & & & & & \\
    GPT5-mini  & -- & -- & 54.7 & 84.4 & 69.5 & 78.6 & 65.3 & 63.2 & 64.2 \\
    GPT-4o~\citep{hurst2024gpt}  & -- & -- & 54.4 & 85.7 & 70.5 & 73.9 & 63.8 & 61.4 & 62.6 \\
    Gemini2.5-Pro  & -- & -- & 59.8 & 89.7 & 74.7 & 76.7 & 69.8 & 63.7 & 66.7 \\
    Claude-4-Sonnet & -- & -- & 53.5 & 90.2 & 71.9 & 45.0 & 63.9 & 59.2 & 61.5 \\
    Qwen2.5-VL~\citep{bai2025qwen25vltechnicalreport} & 72B & 74.0 & 70.4 & 86.7 & 77.0 & 76.9 & 72.1 & 70.0 & 71.1 \\
    \midrule
    \rowcolor[HTML]{F2F2F2} \textit{Peer Models} & & & & & & & & & \\
    Qwen2.5-VL & 3B & 50.1 & 51.6 & 69.2 & 57.0 & 63.3 & 59.2 & 56.1 & 57.7 \\
    Qwen2.5-VL & 7B & 63.0 & 67.7 & 83.1 & 71.3 & 75.9 & 68.3 & 66.0 & 67.2 \\
    Sequential (SFT) & 3B & 51.5 & 52.1 & 70.4 & 58.0 & 65.0 & 60.0 & 58.3 & 59.2 \\
    Sequential (SFT) & 7B & 65.5 & 68.0 & 83.7 & 72.4 & 76.1 & 68.7 & 66.7 & 67.7 \\
    GRPO & 3B & 53.0 & 52.9 & 71.2 & 59.0 & 70.4 & 62.1 & 60.4 & 64.3 \\
    GRPO & 7B & 64.4 & 69.2 & 84.2 & 72.6 & 78.4 & 70.9 & 68.8 & 72.7 \\
    Majority voting@4 & 3B & 52.1 & 52.8 & 69.8 & 58.2 & 64.7 & 60.1 & 58.1 & 59.1 \\
    Majority voting@4 & 7B & 63.2 & 67.6 & 83.2 & 71.3 & 75.6 & 68.9 & 66.8 & 67.8 \\
    \midrule
    Visual Para-Thinker & 3B & 55.0 & 54.4 & 72.6 & 60.7 & 75.9 & 63.6 & 62.2 & 67.2 \\
    Visual Para-Thinker & 7B & 65.7 & 70.8 & 85.4 & 74.0 & 82.2 & 71.3 & 71.0 & 74.8 \\
    \midrule
    \textbf{\method} & 3B & \textbf{60.7} & \textbf{69.5} & \textbf{74.8} & \textbf{68.3} & \textbf{80.0} & \textbf{69.5} & \textbf{64.0} & \textbf{71.2} \\
    \textbf{\method} & 7B & \textbf{68.2} & \textbf{74.0} & \textbf{88.3} & \textbf{76.8} & \textbf{85.7} & \textbf{74.9} & \textbf{73.8} & \textbf{78.1} \\
    \bottomrule
  \end{tabular}%
  }
\end{table*}

\begin{table*}[t]
  \centering
  \small
  \caption{Referring-expression grounding accuracy (Acc@0.5, $\%$) on the RefCOCO family. Numbers in the first three columns are reproduced from~\citet{pan2025parathinker}; \method is trained on the same backbone (Qwen2.5-VL-3B) with the same training data. \textbf{Bold} = best in the row.}
  \label{tab:grounding}
  \setlength{\tabcolsep}{4pt}
  \renewcommand{\arraystretch}{1.10}
  \begin{tabular}{l c c c c}
    \toprule
    \textbf{Split} & \textbf{Qwen2.5-VL-3B} & \textbf{Sequential} & \textbf{Visual Para-Thinker (3B)} & \textbf{\method (3B)} \\
    \midrule
    RefCOCO val   & 85.4 & 85.7 & 88.1 & \textbf{90.5} \\
    RefCOCO testA & 86.1 & 87.3 & 89.8 & \textbf{91.6} \\
    RefCOCO testB & 83.0 & 84.2 & 87.7 & \textbf{90.4} \\
    \rowcolor[gray]{0.92} RefCOCO Avg          & 84.8 & 85.7 & 88.5 & \textbf{90.8} \\
    \midrule
    RefCOCO+ val   & 76.5 & 79.2 & 83.5 & \textbf{87.1} \\
    RefCOCO+ testA & 82.4 & 84.0 & 87.2 & \textbf{89.4} \\
    RefCOCO+ testB & 71.4 & 75.9 & 77.0 & \textbf{80.9} \\
    \rowcolor[gray]{0.92} RefCOCO+ Avg         & 76.7 & 79.7 & 82.6 & \textbf{85.8} \\
    \midrule
    RefCOCOg val  & 84.7 & 86.1 & 87.3 & \textbf{89.8} \\
    RefCOCOg test & 83.1 & 84.0 & 86.9 & \textbf{89.8} \\
    \rowcolor[gray]{0.92} RefCOCOg Avg          & 83.9 & 85.0 & 87.1 & \textbf{89.8} \\
    \bottomrule
  \end{tabular}
\end{table*}

\subsection{Implementation Details}
\label{sec:impl}

We instantiate \method on a Qwen2.5-VL-3B~\citep{bai2025qwen25vltechnicalreport} backbone (with a complementary 7B run in Appendix~\ref{app:7b}). Training proceeds in two stages: Stage~1 performs Multi-Agent Capability Injection under the context-isolation mask of Eq.~\ref{eq:sft}, and Stage~2 performs Role-Decoupled Multi-Agent Optimization with on-policy DAPO~\citep{yu2025dapo} using the role-specific advantages of Eq.~\ref{eq:adv}. 

% =====================================================================
\section{Experiments}
\label{sec:exp}

\subsection{Setup}
\label{sec:setup}
\paragraph{Benchmarks.}
We evaluate on six visual benchmarks that stress different reasoning dimensions:  \textbf{V\textsuperscript{\textasteriskcentered}}~\citep{wu2024vstar} for high-resolution visual search, \textbf{CountBench}~\citep{paiss2023countme} and \textbf{Pixmo}~\citep{molmo2024} for precise object counting, \textbf{MMVP}~\citep{tong2024eyeswideshutexploring} for fine-grained perception, \textbf{RefCOCO / RefCOCO+ / RefCOCOg}~\citep{yu2016modeling} for referring-expression grounding, and \textbf{HallusionBench}~\citep{guan2024hallusionbench} for visual hallucination.

\paragraph{Baselines.}
We compare against (i) the base MLLM (Qwen2.5-VL-3B~\citep{bai2025qwen25vltechnicalreport}) with greedy decoding, (ii) the same base model with long CoT, (iii) self-consistency~\citep{wang2023selfconsistency} with $K$ samples at a comparable token budget, (iv) Para-Thinker~\citep{pan2025parathinker} (parallel thinking without role separation), (v) a multi-agent debate~\citep{du2023improving} variant of the same MLLM, and (vi) a scaled-up Qwen2.5-VL-7B variant reported for reference in Appendix~\ref{app:7b}.

\paragraph{Metrics.}
We report task-specific accuracy/AP, total output tokens per example (Tok.), and wall-clock rollout latency.

\subsection{Main Results}
\label{sec:main}

Table~\ref{tab:main} reports results across counting, fine-grained perception, and hallucination tasks. At the 3B scale, \method lifts the perception-and-hallucination average from $57.7$ on the Qwen2.5-VL-3B backbone to $71.2$ ($+13.5$), and outperforms every peer-scale baseline on every benchmark. Long-CoT-style sequential SFT and Majority Voting@4 leave the counting average essentially unchanged ($57.0\!\to\!58.0/58.2$), and on-policy single-advantage RL (GRPO) closes the perception gap but barely moves counting, indicating that the bottleneck is the underlying single-trajectory perceptual commitment rather than inference compute. Para-Thinker, the closest parallel-thinking baseline without role separation, reaches $60.7$ and $67.2$ on the counting and hallucination averages; \method adds another $+7.6$ and $+4.0$ on top of it under the same multi-path inference template, isolating the contribution of role-specific optimisation.

\begin{figure}[t]
  \centering
  \includegraphics[width=\linewidth]{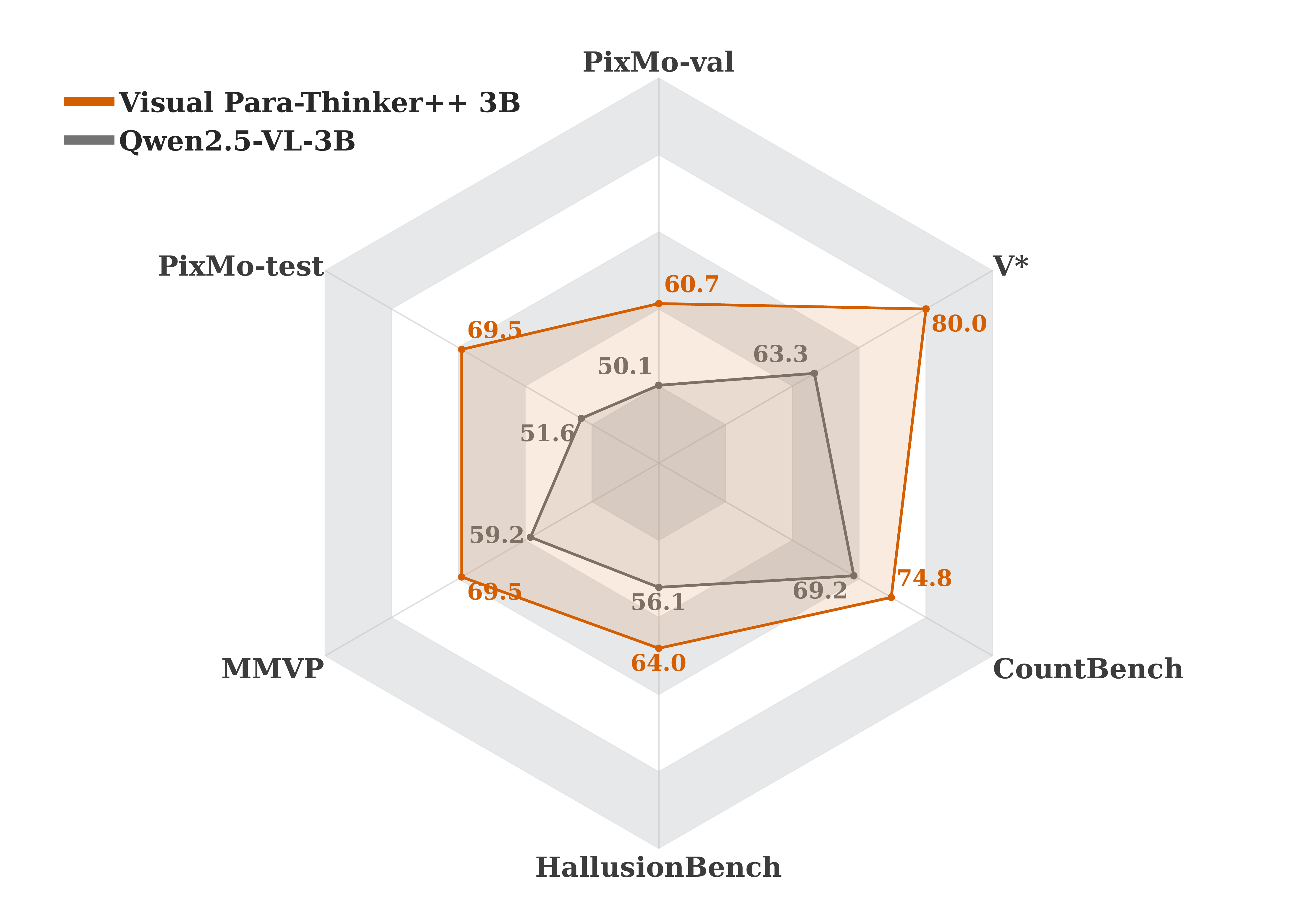}
  \caption{Per-benchmark comparison of \method (3B) against the Qwen2.5-VL-3B backbone on six vision-centric axes drawn directly from Table~\ref{tab:main}.}
  \label{fig:radar}
\end{figure}

The largest gains are on \textbf{Pixmo-test} ($+17.9$), \textbf{V*} ($+16.7$), and \textbf{HallusionBench} ($+7.9$): exactly the benchmarks where the dominant error is early commitment to a wrong region or a hallucinated object. Counting (CountBench) and fine-grained verification (MMVP) show smaller but consistent gains ($+5.6$ and $+10.3$), consistent with the Scan-order pattern reducing single-direction missing-instance errors. Figure~\ref{fig:radar} visualises this margin on six vision-centric axes. The qualitative conclusions transfer to the 7B backbone (Appendix~\ref{app:7b}), where \method-7B remains competitive with frontier closed-source models on V*; the consistent margins across two scales suggest that the design is orthogonal to backbone size.

\begin{table}[t]
  \centering
  \caption{Efficiency comparison on the V\textsuperscript{*} benchmark. ``Majority Voting'' generates four trajectories per query. ``Ours (w/o reuse)'' and ``Ours (w/ reuse)'' refer to our multi-agent rollout engine (\S\ref{sec:engine}) without and with KV-cache reuse across role segments.}
  \label{tab:efficiency}
  \renewcommand{\arraystretch}{1.1}
  \footnotesize
  \setlength{\tabcolsep}{3pt}
  \begin{tabularx}{\columnwidth}{X c c}
    \toprule
    \textbf{Method} & \textbf{Time (s)} $\downarrow$ & \textbf{Throughput (token/s)} $\uparrow$ \\
    \midrule
    Qwen2.5-VL-3B              & 266  & 58  \\
    Sequential                 & 481  & 53  \\
    Majority Voting            & 1197 & 49  \\
    Ours (w/o reuse)           & 364  & 104 \\
    Ours (w/ reuse)            & 312  & 123 \\
    \bottomrule
  \end{tabularx}
\end{table}

\begin{table*}[t]
  \centering
  \small
  \caption{Cross-paradigm comparison of reward composition on the Qwen2.5-VL-3B backbone using an identical Worker-reward signal $R_w$ and the same Stage-1 checkpoint. \textbf{Bold} = best.}
  \label{tab:ablation-composition}
  \vspace{-3pt}
  \setlength{\tabcolsep}{6pt}
  \renewcommand{\arraystretch}{1.25}
  \begin{tabular}{l|c|c|c|c|c}
    \toprule
    \textbf{Composition} & \textbf{Reward $\to$ advantage} & \textbf{CountBench} & \textbf{V*} & \textbf{MMVP} & \textbf{HallusionBench} \\
    \midrule
    Naive Sum         & $R_o{+}{\textstyle\sum_i}R_w^{(i)} \to A_\text{shared}$  & 58.7          & 66.4          & 61.3          & 50.9          \\
    Conditional Sum   & $R_o{+}R_o\!\cdot\!{\textstyle\sum_i}R_w^{(i)} \to A_\text{shared}$        & 62.9          & 69.8          & 65.1          & 55.2          \\
    Role-decoupled $A$ (\textbf{Ours}) & $R_o \to A_\text{out};\ R_w^{(i)} \to A_w^{(i)}$ & \textbf{66.2} & \textbf{72.8} & \textbf{68.7} & \textbf{59.0} \\
    \bottomrule
  \end{tabular}
\end{table*}

\begin{table*}[t]
  \centering
  \small
  \caption{Factorial ablation of the two design switches in Role-Decoupled Multi-Agent Optimization (\S\ref{sec:rdmao}). \textbf{$R_w$ (Worker reward)}: \cmark\ enables the per-Worker local-quality signal $R_\text{worker}^{(i)}$ on top of the outcome reward; \xmark\ uses outcome reward only. \textbf{Role-decoupled $A$ (advantage)}: \cmark\ assigns role-specific advantages following Eq.~\ref{eq:adv} ($A_\text{out}$ for Main/Summary tokens; $A_\text{out}{+}\lambda A_\text{worker}^{(i)}$ for Worker $i$'s tokens); \xmark\ uses a single shared advantage on every token.}
  \label{tab:ablation-reward}
  \vspace{-3pt}
  \setlength{\tabcolsep}{6pt}
  \renewcommand{\arraystretch}{1.15}
  \begin{tabular}{c|c|c|c|c|c}
    \toprule
    \textbf{Worker reward} & \textbf{Role-decoupled} & \multirow{2}{*}{\textbf{CountBench}} & \multirow{2}{*}{\textbf{V*}} & \multirow{2}{*}{\textbf{MMVP}} & \multirow{2}{*}{\textbf{HallusionBench}} \\
    $R_w$  & advantage $A$ & & & & \\
    \midrule
    \xmark & \xmark & 60.4          & 68.1          & 62.7          & 52.7          \\
    \cmark & \xmark & 63.5          & 70.6          & 65.9          & 55.8          \\
    \xmark & \cmark & 64.7          & 71.2          & 66.8          & 56.9          \\
    \cmark & \cmark & \textbf{66.2} & \textbf{72.8} & \textbf{68.7} & \textbf{59.0} \\
    \bottomrule
  \end{tabular}
\end{table*}

\paragraph{Grounding (RefCOCO family).}
We further evaluate referring-expression grounding on the RefCOCO/+/g splits (Acc@0.5; Table~\ref{tab:grounding}). \method outperforms both the Qwen2.5-VL-3B backbone and the SFT-aligned Visual Para-Thinker on every split. The improvement over Visual Para-Thinker is consistent across all nine entries (between $+1.8$ and $+3.9$ points), with the largest residual gain on the harder RefCOCO+ splits (val $+3.6$, testB $+3.9$), where attribute-and-relation cues---rather than pure spatial cues---play a larger role. This is consistent with the picture from \S\ref{sec:main}: when the dominant error mode is early commitment to an incorrect region or attribute, role-decoupled optimisation over four parallel Worker reasoning paths reconciled by the Summary Agent yields gains that single-trajectory CoT and shared-advantage RL cannot reach.

\subsection{Ablation Studies}
\label{sec:ablation}

\paragraph{How should $R_w$ and $R_o$ be composed?}
Table~\ref{tab:ablation-composition} compares the three reward-composition paradigms of Fig.~\ref{fig:reward} under an identical Worker-reward signal $R_w$. Naive Sum is the worst of the three and falls \emph{below} the outcome-only baseline ($R_w$ off, see the (\xmark, \xmark) row of Table~\ref{tab:ablation-reward}), confirming the gradient-coupling failure mode of \S\ref{sec:rdmao}. Conditional Sum recovers most of this loss but is still bounded by the shared-advantage bottleneck. Composing the two rewards at the \emph{advantage} level, as in our role-decoupled scheme, yields the largest and most consistent gains across all four benchmarks.

% \begin{table*}[t]
%   ... duplicate of tab:ablation-reward, kept only as a backup of caption text ...
% \end{table*}

\paragraph{Are $R_w$ and role-decoupled $A$ each necessary?}
Table~\ref{tab:ablation-reward} factorises Role-Decoupled Multi-Agent Optimization into two orthogonal switches: Worker reward $R_w$ on/off and role-decoupled advantage on/off. Role-decoupled advantage alone (\xmark, \cmark) already lifts CountBench by $+4.3$ and HallusionBench by $+4.2$ over the (\xmark, \xmark) baseline, indicating that gradient routing across role segments is the dominant lever; adding $R_w$ on top contributes a further $+1.5$ to $+2.1$, with the largest residual gain on MMVP. Adding $R_w$ on a shared advantage (\cmark, \xmark) yields a smaller and less consistent gain, reproducing the gradient-coupling pattern of Naive Sum. The two design choices are therefore complementary: $R_w$ supplies the local signal, and role-decoupled $A$ keeps it from polluting the Main and Summary gradients.

\paragraph{Sensitivity to the Worker advantage weight $\lambda$.}
Setting $\lambda{=}0$ in Eq.~\ref{eq:adv} recovers an outcome-only RL run equivalent to GRPO over the multi-agent trajectory, trailing the full method by $3.1$--$4.4$ points across V*, CountBench, and HallusionBench. Increasing $\lambda$ from $0$ to $0.5$ monotonically improves all three; pushing $\lambda$ to $1.0$ or $2.0$ over-weights the noisy Worker reward and partially erases the gain. We use $\lambda{=}0.5$ throughout (full sweep in Appendix~\ref{app:abl}, Table~\ref{tab:lambda}).

\paragraph{Effect of the number of Worker Agents $K$.}
Reducing $K$ from $4$ to $2$ drops V$^{*}$ by $2.7$ and CountBench by $3.1$, while increasing $K$ to $8$ yields negligible gains ($+0.3$ on average) at nearly doubled rollout cost, confirming $K{=}4$ as the accuracy--efficiency sweet spot under our fixed allocation protocol.

\subsection{Efficiency Analysis}
\label{sec:efficiency}

Table~\ref{tab:efficiency} reports the inference efficiency on the V\textsuperscript{*} benchmark following the Visual Para-Thinker evaluation setting. With KV-cache reuse across role segments, \method keeps total inference time close to the base Qwen2.5-VL-3B model while substantially improving throughput over sequential and majority-voting parallel baselines.

% =====================================================================
\section{Conclusion}
\label{sec:conclusion}

We presented \method, a single-policy multi-agent framework that instantiates one shared MLLM as Main, Worker, and Summary Agents under a fixed protocol, trained with Multi-Agent Capability Injection and Role-Decoupled Multi-Agent Optimization to reduce reward conflict across roles. Across six benchmarks, \method delivers consistent gains over single-trajectory and parallel baselines.

% =====================================================================
\section*{Limitations}
\label{sec:limitations}

Our study has several limitations. First, our primary experiments are conducted on a Qwen2.5-VL-3B backbone, with a complementary 7B run reported for reference in Appendix~\ref{app:7b}; the scaling behaviour on substantially larger backbones, such as models above 30B parameters, and on alternative role organizations remains to be characterized. Second, the empirical study covers six benchmarks that emphasize visual search, counting, fine-grained perception, grounding, and hallucination detection, but does not yet examine broader multimodal settings such as document understanding, chart reasoning, or video-based decision making. Third, our current Main Agent uses a fixed task-allocation protocol with two task-dependent patterns: Block-based allocation and Scan-order allocation. We deliberately do not model the Main Agent as a free-form open-ended planner in this paper. Adaptive task allocation, including when to switch patterns, when to vary the number of parallel Worker reasoning paths, and how to compose more granular Worker roles, remains an important direction for future work. Fourth, although the native multi-agent inference engine reduces redundant prefill and KV cache overhead, horizontal scaling still increases overall decoding cost as additional Worker reasoning paths are introduced; in our 3B experiments the four-path configuration provides the best accuracy-cost trade-off, and this trade-off may shift for larger backbones. Fifth, the Worker reward is currently instantiated by an agreement-driven majority-vote heuristic, which can be brittle when most Worker reasoning paths share the same hallucination; the Summary reward is currently instantiated by the final-answer outcome reward, and we do not yet explicitly supervise the faithfulness of the Summary Agent's reconciliation. Stronger process supervision for both Worker and Summary roles is a natural extension. Finally, we do not yet isolate the Summary Agent's decision policy with a dedicated ablation; future work should separate the contribution of trace reading from that of evidence reconciliation, for instance by replacing the trained Summary segment with a non-learned majority vote at inference time. Addressing adaptive Main-Agent task allocation, stronger process supervision for Worker and Summary rewards, Summary-Agent decision-policy ablation, and broader evaluation across multimodal domains remain future work.

% =====================================================================
\section*{Ethics Statement}
\label{sec:ethics}

Our work aims to improve the reliability of multimodal reasoning systems by encouraging explicit exploration of multiple visual hypotheses through role-conditioned Worker Agents before producing a final answer. In principle, this can have positive downstream effects in settings where premature commitment to a single interpretation is costly, such as assistive interfaces, educational tools, and decision-support systems that rely on visual evidence. The native multi-agent inference engine is also designed to reduce computational cost relative to naive role-separated rollout through KV cache reuse, which may modestly reduce the marginal cost of multi-agent reasoning.

At the same time, improved visual reasoning can also amplify misuse risks. More capable multimodal models may be used to generate more persuasive but incorrect analyses of images, documents, or screenshots, including misleading explanations that appear internally coherent. Moreover, our method does not eliminate hallucinations; it only reduces them empirically on the evaluated benchmarks. In high-stakes applications, a more confident multi-agent system could still produce incorrect outputs, potentially with a stronger veneer of justification. For this reason, we do not view \method as a substitute for domain-specific verification, human oversight, or calibrated uncertainty estimation.

The training pipeline also raises practical and methodological considerations. First, the Worker reward used in Stage~2 relies on agreement among intermediate Worker traces, which may reinforce majority errors when all Worker reasoning paths share the same blind spot. Second, teacher-generated SFT traces may encode biases or stylistic artifacts from the upstream teacher system. Third, multi-agent reasoning increases total generated tokens even when inference is optimized, which has compute and environmental costs. Throughout this paper we therefore report both accuracy and inference cost, evaluate explicitly on hallucination-sensitive benchmarks, and caution against deploying \method in safety-critical scenarios without additional safeguards and external verification mechanisms.

% =====================================================================
% Acknowledgments are omitted in the anonymous submission version.
% \section*{Acknowledgments}

% \bibliographystyle{acl}

\bibliography{acl}

@inproceedings{wei2022chain,
      title={Chain-of-Thought Prompting Elicits Reasoning in Large Language Models}, 
      author={Jason Wei and Xuezhi Wang and Dale Schuurmans and Maarten Bosma and Brian Ichter and Fei Xia and Ed Chi and Quoc Le and Denny Zhou},
      year={2023},
      eprint={2201.11903},
      archivePrefix={arXiv},
      primaryClass={cs.CL},
      url={https://arxiv.org/abs/2201.11903}, 
}

@misc{liu2025efficientreasoningsuppressionselfaffirmation,
      title={Efficient Reasoning Through Suppression of Self-Affirmation Reflections in Large Reasoning Models}, 
      author={Kaiyuan Liu and Chen Shen and Zhanwei Zhang and Junjie Liu and Xiaosong Yuan and Jieping ye},
      year={2025},
      eprint={2506.12353},
      archivePrefix={arXiv},
      primaryClass={cs.CL},
      url={https://arxiv.org/abs/2506.12353}, 
}

@misc{openai2024o1,
      title={OpenAI o1 System Card}, 
      author={OpenAI and : and Aaron Jaech and Adam Kalai and Adam Lerer and Adam Richardson and Ahmed El-Kishky and Aiden Low and Alec Helyar and Aleksander Madry and Alex Beutel and Alex Carney and Alex Iftimie and Alex Karpenko and Alex Tachard Passos and Alexander Neitz and Alexander Prokofiev and Alexander Wei and Allison Tam and Ally Bennett and Ananya Kumar and Andre Saraiva and Andrea Vallone and Andrew Duberstein and Andrew Kondrich and Andrey Mishchenko and Andy Applebaum and Angela Jiang and Ashvin Nair and Barret Zoph and Behrooz Ghorbani and Bohan Zhang and Ben Rossen and Benjamin Sokolowsky and Boaz Barak and Bob McGrew and Borys Minaiev and Botao Hao and Bowen Baker and Brandon Houghton and Brandon McKinzie and Brydon Eastman and Camillo Lugaresi and Cary Bassin and Cary Hudson and Chak Ming Li and Charles de Bourcy and Chelsea Voss and Chen Shen and Chong Zhang and Chris Koch and Chris Orsinger and Christopher Hesse and Claudia Fischer and Clive Chan and Dan Roberts and Daniel Kappler and Daniel Levy and Daniel Selsam and David Dohan and David Farhi and David Mely and David Robinson and Dimitris Tsipras and Doug Li and Dragos Oprica and Eben Freeman and Eddie Zhang and Edmund Wong and Elizabeth Proehl and Enoch Cheung and Eric Mitchell and Eric Wallace and Erik Ritter and Evan Mays and Fan Wang and Felipe Petroski Such and Filippo Raso and Florencia Leoni and Foivos Tsimpourlas and Francis Song and Fred von Lohmann and Freddie Sulit and Geoff Salmon and Giambattista Parascandolo and Gildas Chabot and Grace Zhao and Greg Brockman and Guillaume Leclerc and Hadi Salman and Haiming Bao and Hao Sheng and Hart Andrin and Hessam Bagherinezhad and Hongyu Ren and Hunter Lightman and Hyung Won Chung and Ian Kivlichan and Ian O'Connell and Ian Osband and Ignasi Clavera Gilaberte and Ilge Akkaya and Ilya Kostrikov and Ilya Sutskever and Irina Kofman and Jakub Pachocki and James Lennon and Jason Wei and Jean Harb and Jerry Twore and Jiacheng Feng and Jiahui Yu and Jiayi Weng and Jie Tang and Jieqi Yu and Joaquin Quiñonero Candela and Joe Palermo and Joel Parish and Johannes Heidecke and John Hallman and John Rizzo and Jonathan Gordon and Jonathan Uesato and Jonathan Ward and Joost Huizinga and Julie Wang and Kai Chen and Kai Xiao and Karan Singhal and Karina Nguyen and Karl Cobbe and Katy Shi and Kayla Wood and Kendra Rimbach and Keren Gu-Lemberg and Kevin Liu and Kevin Lu and Kevin Stone and Kevin Yu and Lama Ahmad and Lauren Yang and Leo Liu and Leon Maksin and Leyton Ho and Liam Fedus and Lilian Weng and Linden Li and Lindsay McCallum and Lindsey Held and Lorenz Kuhn and Lukas Kondraciuk and Lukasz Kaiser and Luke Metz and Madelaine Boyd and Maja Trebacz and Manas Joglekar and Mark Chen and Marko Tintor and Mason Meyer and Matt Jones and Matt Kaufer and Max Schwarzer and Meghan Shah and Mehmet Yatbaz and Melody Y. Guan and Mengyuan Xu and Mengyuan Yan and Mia Glaese and Mianna Chen and Michael Lampe and Michael Malek and Michele Wang and Michelle Fradin and Mike McClay and Mikhail Pavlov and Miles Wang and Mingxuan Wang and Mira Murati and Mo Bavarian and Mostafa Rohaninejad and Nat McAleese and Neil Chowdhury and Neil Chowdhury and Nick Ryder and Nikolas Tezak and Noam Brown and Ofir Nachum and Oleg Boiko and Oleg Murk and Olivia Watkins and Patrick Chao and Paul Ashbourne and Pavel Izmailov and Peter Zhokhov and Rachel Dias and Rahul Arora and Randall Lin and Rapha Gontijo Lopes and Raz Gaon and Reah Miyara and Reimar Leike and Renny Hwang and Rhythm Garg and Robin Brown and Roshan James and Rui Shu and Ryan Cheu and Ryan Greene and Saachi Jain and Sam Altman and Sam Toizer and Sam Toyer and Samuel Miserendino and Sandhini Agarwal and Santiago Hernandez and Sasha Baker and Scott McKinney and Scottie Yan and Shengjia Zhao and Shengli Hu and Shibani Santurkar and Shraman Ray Chaudhuri and Shuyuan Zhang and Siyuan Fu and Spencer Papay and Steph Lin and Suchir Balaji and Suvansh Sanjeev and Szymon Sidor and Tal Broda and Aidan Clark and Tao Wang and Taylor Gordon and Ted Sanders and Tejal Patwardhan and Thibault Sottiaux and Thomas Degry and Thomas Dimson and Tianhao Zheng and Timur Garipov and Tom Stasi and Trapit Bansal and Trevor Creech and Troy Peterson and Tyna Eloundou and Valerie Qi and Vineet Kosaraju and Vinnie Monaco and Vitchyr Pong and Vlad Fomenko and Weiyi Zheng and Wenda Zhou and Wenting Zhan and Wes McCabe and Wojciech Zaremba and Yann Dubois and Yinghai Lu and Yining Chen and Young Cha and Yu Bai and Yuchen He and Yuchen Zhang and Yunyun Wang and Zheng Shao and Zhuohan Li},
      year={2026},
      eprint={2412.16720},
      archivePrefix={arXiv},
      primaryClass={cs.AI},
      url={https://arxiv.org/abs/2412.16720}, 
}

@inproceedings{wang2023selfconsistency,
      title={Self-Consistency Improves Chain of Thought Reasoning in Language Models}, 
      author={Xuezhi Wang and Jason Wei and Dale Schuurmans and Quoc Le and Ed Chi and Sharan Narang and Aakanksha Chowdhery and Denny Zhou},
      year={2023},
      eprint={2203.11171},
      archivePrefix={arXiv},
      primaryClass={cs.CL},
      url={https://arxiv.org/abs/2203.11171}, 
}

@inproceedings{yao2023tree,
      title={Tree of Thoughts: Deliberate Problem Solving with Large Language Models}, 
      author={Shunyu Yao and Dian Yu and Jeffrey Zhao and Izhak Shafran and Thomas L. Griffiths and Yuan Cao and Karthik Narasimhan},
      year={2023},
      eprint={2305.10601},
      archivePrefix={arXiv},
      primaryClass={cs.CL},
      url={https://arxiv.org/abs/2305.10601}, 
}

@misc{du2023improving,
      title={Improving Factuality and Reasoning in Language Models through Multiagent Debate}, 
      author={Yilun Du and Shuang Li and Antonio Torralba and Joshua B. Tenenbaum and Igor Mordatch},
      year={2023},
      eprint={2305.14325},
      archivePrefix={arXiv},
      primaryClass={cs.CL},
      url={https://arxiv.org/abs/2305.14325}, 
}

@article{chen2023agentverse,
      title={AgentVerse: Facilitating Multi-Agent Collaboration and Exploring Emergent Behaviors}, 
      author={Weize Chen and Yusheng Su and Jingwei Zuo and Cheng Yang and Chenfei Yuan and Chi-Min Chan and Heyang Yu and Yaxi Lu and Yi-Hsin Hung and Chen Qian and Yujia Qin and Xin Cong and Ruobing Xie and Zhiyuan Liu and Maosong Sun and Jie Zhou},
      year={2023},
      eprint={2308.10848},
      archivePrefix={arXiv},
      primaryClass={cs.CL},
      url={https://arxiv.org/abs/2308.10848}, 
}

@inproceedings{liu2024llava,
        title={Visual Instruction Tuning}, 
      author={Haotian Liu and Chunyuan Li and Qingyang Wu and Yong Jae Lee},
      year={2023},
      eprint={2304.08485},
      archivePrefix={arXiv},
      primaryClass={cs.CV},
      url={https://arxiv.org/abs/2304.08485}, 
}

@article{zhang2023multimodal,
      title={Multimodal Chain-of-Thought Reasoning in Language Models}, 
      author={Zhuosheng Zhang and Aston Zhang and Mu Li and Hai Zhao and George Karypis and Alex Smola},
      year={2024},
      eprint={2302.00923},
      archivePrefix={arXiv},
      primaryClass={cs.CL},
      url={https://arxiv.org/abs/2302.00923}, 
}

@inproceedings{mitra2024compositional,
      title={Compositional Chain-of-Thought Prompting for Large Multimodal Models}, 
      author={Chancharik Mitra and Brandon Huang and Trevor Darrell and Roei Herzig},
      year={2024},
      eprint={2311.17076},
      archivePrefix={arXiv},
      primaryClass={cs.CV},
      url={https://arxiv.org/abs/2311.17076}, 
}

@inproceedings{wu2024vstar,
      title={V*: Guided Visual Search as a Core Mechanism in Multimodal LLMs}, 
      author={Penghao Wu and Saining Xie},
      year={2023},
      eprint={2312.14135},
      archivePrefix={arXiv},
      primaryClass={cs.CV},
      url={https://arxiv.org/abs/2312.14135}, 
}

@inproceedings{paiss2023countme,
      title={Teaching CLIP to Count to Ten}, 
      author={Roni Paiss and Ariel Ephrat and Omer Tov and Shiran Zada and Inbar Mosseri and Michal Irani and Tali Dekel},
      year={2023},
      eprint={2302.12066},
      archivePrefix={arXiv},
      primaryClass={cs.CV},
      url={https://arxiv.org/abs/2302.12066}, 
}

@inproceedings{yu2016modeling,
      title={Modeling Context in Referring Expressions}, 
      author={Licheng Yu and Patrick Poirson and Shan Yang and Alexander C. Berg and Tamara L. Berg},
      year={2016},
      eprint={1608.00272},
      archivePrefix={arXiv},
      primaryClass={cs.CV},
      url={https://arxiv.org/abs/1608.00272}, 
}

@inproceedings{guan2024hallusionbench,
  title={HallusionBench: An Advanced Diagnostic Suite for Entangled Language Hallucination and Visual Illusion in Large Vision-Language Models}, 
      author={Tianrui Guan and Fuxiao Liu and Xiyang Wu and Ruiqi Xian and Zongxia Li and Xiaoyu Liu and Xijun Wang and Lichang Chen and Furong Huang and Yaser Yacoob and Dinesh Manocha and Tianyi Zhou},
      year={2024},
      eprint={2310.14566},
      archivePrefix={arXiv},
      primaryClass={cs.CV},
      url={https://arxiv.org/abs/2310.14566}, 
}

@article{yu2025dapo,
  title={DAPO: An Open-Source LLM Reinforcement Learning System at Scale}, 
      author={Qiying Yu and Zheng Zhang and Ruofei Zhu and Yufeng Yuan and Xiaochen Zuo and Yu Yue and Weinan Dai and Tiantian Fan and Gaohong Liu and Lingjun Liu and Xin Liu and Haibin Lin and Zhiqi Lin and Bole Ma and Guangming Sheng and Yuxuan Tong and Chi Zhang and Mofan Zhang and Wang Zhang and Hang Zhu and Jinhua Zhu and Jiaze Chen and Jiangjie Chen and Chengyi Wang and Hongli Yu and Yuxuan Song and Xiangpeng Wei and Hao Zhou and Jingjing Liu and Wei-Ying Ma and Ya-Qin Zhang and Lin Yan and Mu Qiao and Yonghui Wu and Mingxuan Wang},
      year={2025},
      eprint={2503.14476},
      archivePrefix={arXiv},
      primaryClass={cs.LG},
      url={https://arxiv.org/abs/2503.14476}, 
}

@article{nvidia2026gdpo,
  title={GDPO: Group reward-Decoupled Normalization Policy Optimization for Multi-reward RL Optimization}, 
      author={Shih-Yang Liu and Xin Dong and Ximing Lu and Shizhe Diao and Peter Belcak and Mingjie Liu and Min-Hung Chen and Hongxu Yin and Yu-Chiang Frank Wang and Kwang-Ting Cheng and Yejin Choi and Jan Kautz and Pavlo Molchanov},
      year={2026},
      eprint={2601.05242},
      archivePrefix={arXiv},
      primaryClass={cs.CL},
      url={https://arxiv.org/abs/2601.05242}, 
}

@inproceedings{
yuan2026differential,
title={Differential Fine-Tuning Large Language Models Towards Better Diverse Reasoning Abilities},
author={Xiaosong Yuan and Chen Shen and Shaotian Yan and kaiyuan liu and Xiaofeng Zhang and Sinan Fan and Liang Xie and Wenxiao Wang and Renchu Guan and Ying Wang and Jieping Ye},
booktitle={The Fourteenth International Conference on Learning Representations},
year={2026},
url={https://openreview.net/forum?id=aIhn4GhTBW}
}

@article{xu2024llavao1,
  title={Direct Preference Optimization: Your Language Model is Secretly a Reward Model}, 
      author={Rafael Rafailov and Archit Sharma and Eric Mitchell and Stefano Ermon and Christopher D. Manning and Chelsea Finn},
      year={2024},
      eprint={2305.18290},
      archivePrefix={arXiv},
      primaryClass={cs.LG},
      url={https://arxiv.org/abs/2305.18290}, 
}

@inproceedings{shao2024visualcot,
  title={Visual CoT: Advancing Multi-Modal Language Models with a Comprehensive Dataset and Benchmark for Chain-of-Thought Reasoning}, 
      author={Hao Shao and Shengju Qian and Han Xiao and Guanglu Song and Zhuofan Zong and Letian Wang and Yu Liu and Hongsheng Li},
      year={2024},
      eprint={2403.16999},
      archivePrefix={arXiv},
      primaryClass={cs.CV},
      url={https://arxiv.org/abs/2403.16999}, 
}

@misc{pan2025parathinker,
  title={ParaThinker: Native Parallel Thinking as a New Paradigm to Scale LLM Test-time Compute}, 
      author={Hao Wen and Yifan Su and Feifei Zhang and Yunxin Liu and Yunhao Liu and Ya-Qin Zhang and Yuanchun Li},
      year={2025},
      eprint={2509.04475},
      archivePrefix={arXiv},
      primaryClass={cs.CL},
      url={https://arxiv.org/abs/2509.04475}, 
}

@article{hurst2024gpt,
  title={GPT-4o System Card}, 
      author={OpenAI and : and Aaron Hurst and Adam Lerer and Adam P. Goucher and Adam Perelman and Aditya Ramesh and Aidan Clark and AJ Ostrow and Akila Welihinda and Alan Hayes and Alec Radford and Aleksander Mądry and Alex Baker-Whitcomb and Alex Beutel and Alex Borzunov and Alex Carney and Alex Chow and Alex Kirillov and Alex Nichol and Alex Paino and Alex Renzin and Alex Tachard Passos and Alexander Kirillov and Alexi Christakis and Alexis Conneau and Ali Kamali and Allan Jabri and Allison Moyer and Allison Tam and Amadou Crookes and Amin Tootoochian and Amin Tootoonchian and Ananya Kumar and Andrea Vallone and Andrej Karpathy and Andrew Braunstein and Andrew Cann and Andrew Codispoti and Andrew Galu and Andrew Kondrich and Andrew Tulloch and Andrey Mishchenko and Angela Baek and Angela Jiang and Antoine Pelisse and Antonia Woodford and Anuj Gosalia and Arka Dhar and Ashley Pantuliano and Avi Nayak and Avital Oliver and Barret Zoph and Behrooz Ghorbani and Ben Leimberger and Ben Rossen and Ben Sokolowsky and Ben Wang and Benjamin Zweig and Beth Hoover and Blake Samic and Bob McGrew and Bobby Spero and Bogo Giertler and Bowen Cheng and Brad Lightcap and Brandon Walkin and Brendan Quinn and Brian Guarraci and Brian Hsu and Bright Kellogg and Brydon Eastman and Camillo Lugaresi and Carroll Wainwright and Cary Bassin and Cary Hudson and Casey Chu and Chad Nelson and Chak Li and Chan Jun Shern and Channing Conger and Charlotte Barette and Chelsea Voss and Chen Ding and Cheng Lu and Chong Zhang and Chris Beaumont and Chris Hallacy and Chris Koch and Christian Gibson and Christina Kim and Christine Choi and Christine McLeavey and Christopher Hesse and Claudia Fischer and Clemens Winter and Coley Czarnecki and Colin Jarvis and Colin Wei and Constantin Koumouzelis and Dane Sherburn and Daniel Kappler and Daniel Levin and Daniel Levy and David Carr and David Farhi and David Mely and David Robinson and David Sasaki and Denny Jin and Dev Valladares and Dimitris Tsipras and Doug Li and Duc Phong Nguyen and Duncan Findlay and Edede Oiwoh and Edmund Wong and Ehsan Asdar and Elizabeth Proehl and Elizabeth Yang and Eric Antonow and Eric Kramer and Eric Peterson and Eric Sigler and Eric Wallace and Eugene Brevdo and Evan Mays and Farzad Khorasani and Felipe Petroski Such and Filippo Raso and Francis Zhang and Fred von Lohmann and Freddie Sulit and Gabriel Goh and Gene Oden and Geoff Salmon and Giulio Starace and Greg Brockman and Hadi Salman and Haiming Bao and Haitang Hu and Hannah Wong and Haoyu Wang and Heather Schmidt and Heather Whitney and Heewoo Jun and Hendrik Kirchner and Henrique Ponde de Oliveira Pinto and Hongyu Ren and Huiwen Chang and Hyung Won Chung and Ian Kivlichan and Ian O'Connell and Ian O'Connell and Ian Osband and Ian Silber and Ian Sohl and Ibrahim Okuyucu and Ikai Lan and Ilya Kostrikov and Ilya Sutskever and Ingmar Kanitscheider and Ishaan Gulrajani and Jacob Coxon and Jacob Menick and Jakub Pachocki and James Aung and James Betker and James Crooks and James Lennon and Jamie Kiros and Jan Leike and Jane Park and Jason Kwon and Jason Phang and Jason Teplitz and Jason Wei and Jason Wolfe and Jay Chen and Jeff Harris and Jenia Varavva and Jessica Gan Lee and Jessica Shieh and Ji Lin and Jiahui Yu and Jiayi Weng and Jie Tang and Jieqi Yu and Joanne Jang and Joaquin Quinonero Candela and Joe Beutler and Joe Landers and Joel Parish and Johannes Heidecke and John Schulman and Jonathan Lachman and Jonathan McKay and Jonathan Uesato and Jonathan Ward and Jong Wook Kim and Joost Huizinga and Jordan Sitkin and Jos Kraaijeveld and Josh Gross and Josh Kaplan and Josh Snyder and Joshua Achiam and Joy Jiao and Joyce Lee and Juntang Zhuang and Justyn Harriman and Kai Fricke and Kai Hayashi and Karan Singhal and Katy Shi and Kavin Karthik and Kayla Wood and Kendra Rimbach and Kenny Hsu and Kenny Nguyen and Keren Gu-Lemberg and Kevin Button and Kevin Liu and Kiel Howe and Krithika Muthukumar and Kyle Luther and Lama Ahmad and Larry Kai and Lauren Itow and Lauren Workman and Leher Pathak and Leo Chen and Li Jing and Lia Guy and Liam Fedus and Liang Zhou and Lien Mamitsuka and Lilian Weng and Lindsay McCallum and Lindsey Held and Long Ouyang and Louis Feuvrier and Lu Zhang and Lukas Kondraciuk and Lukasz Kaiser and Luke Hewitt and Luke Metz and Lyric Doshi and Mada Aflak and Maddie Simens and Madelaine Boyd and Madeleine Thompson and Marat Dukhan and Mark Chen and Mark Gray and Mark Hudnall and Marvin Zhang and Marwan Aljubeh and Mateusz Litwin and Matthew Zeng and Max Johnson and Maya Shetty and Mayank Gupta and Meghan Shah and Mehmet Yatbaz and Meng Jia Yang and Mengchao Zhong and Mia Glaese and Mianna Chen and Michael Janner and Michael Lampe and Michael Petrov and Michael Wu and Michele Wang and Michelle Fradin and Michelle Pokrass and Miguel Castro and Miguel Oom Temudo de Castro and Mikhail Pavlov and Miles Brundage and Miles Wang and Minal Khan and Mira Murati and Mo Bavarian and Molly Lin and Murat Yesildal and Nacho Soto and Natalia Gimelshein and Natalie Cone and Natalie Staudacher and Natalie Summers and Natan LaFontaine and Neil Chowdhury and Nick Ryder and Nick Stathas and Nick Turley and Nik Tezak and Niko Felix and Nithanth Kudige and Nitish Keskar and Noah Deutsch and Noel Bundick and Nora Puckett and Ofir Nachum and Ola Okelola and Oleg Boiko and Oleg Murk and Oliver Jaffe and Olivia Watkins and Olivier Godement and Owen Campbell-Moore and Patrick Chao and Paul McMillan and Pavel Belov and Peng Su and Peter Bak and Peter Bakkum and Peter Deng and Peter Dolan and Peter Hoeschele and Peter Welinder and Phil Tillet and Philip Pronin and Philippe Tillet and Prafulla Dhariwal and Qiming Yuan and Rachel Dias and Rachel Lim and Rahul Arora and Rajan Troll and Randall Lin and Rapha Gontijo Lopes and Raul Puri and Reah Miyara and Reimar Leike and Renaud Gaubert and Reza Zamani and Ricky Wang and Rob Donnelly and Rob Honsby and Rocky Smith and Rohan Sahai and Rohit Ramchandani and Romain Huet and Rory Carmichael and Rowan Zellers and Roy Chen and Ruby Chen and Ruslan Nigmatullin and Ryan Cheu and Saachi Jain and Sam Altman and Sam Schoenholz and Sam Toizer and Samuel Miserendino and Sandhini Agarwal and Sara Culver and Scott Ethersmith and Scott Gray and Sean Grove and Sean Metzger and Shamez Hermani and Shantanu Jain and Shengjia Zhao and Sherwin Wu and Shino Jomoto and Shirong Wu and Shuaiqi and Xia and Sonia Phene and Spencer Papay and Srinivas Narayanan and Steve Coffey and Steve Lee and Stewart Hall and Suchir Balaji and Tal Broda and Tal Stramer and Tao Xu and Tarun Gogineni and Taya Christianson and Ted Sanders and Tejal Patwardhan and Thomas Cunninghman and Thomas Degry and Thomas Dimson and Thomas Raoux and Thomas Shadwell and Tianhao Zheng and Todd Underwood and Todor Markov and Toki Sherbakov and Tom Rubin and Tom Stasi and Tomer Kaftan and Tristan Heywood and Troy Peterson and Tyce Walters and Tyna Eloundou and Valerie Qi and Veit Moeller and Vinnie Monaco and Vishal Kuo and Vlad Fomenko and Wayne Chang and Weiyi Zheng and Wenda Zhou and Wesam Manassra and Will Sheu and Wojciech Zaremba and Yash Patil and Yilei Qian and Yongjik Kim and Youlong Cheng and Yu Zhang and Yuchen He and Yuchen Zhang and Yujia Jin and Yunxing Dai and Yury Malkov},
      year={2024},
      eprint={2410.21276},
      archivePrefix={arXiv},
      primaryClass={cs.CL},
      url={https://arxiv.org/abs/2410.21276}, 
}

@misc{bai2025qwen25vltechnicalreport,
  title={Qwen2.5-VL Technical Report}, 
      author={Shuai Bai and Keqin Chen and Xuejing Liu and Jialin Wang and Wenbin Ge and Sibo Song and Kai Dang and Peng Wang and Shijie Wang and Jun Tang and Humen Zhong and Yuanzhi Zhu and Mingkun Yang and Zhaohai Li and Jianqiang Wan and Pengfei Wang and Wei Ding and Zheren Fu and Yiheng Xu and Jiabo Ye and Xi Zhang and Tianbao Xie and Zesen Cheng and Hang Zhang and Zhibo Yang and Haiyang Xu and Junyang Lin},
      year={2025},
      eprint={2502.13923},
      archivePrefix={arXiv},
      primaryClass={cs.CV},
      url={https://arxiv.org/abs/2502.13923}, 
}

@inproceedings{ye2025kvcommonlinecrosscontextkvcache,
  title={KVCOMM: Online Cross-context KV-cache Communication for Efficient LLM-based Multi-agent Systems}, 
      author={Hancheng Ye and Zhengqi Gao and Mingyuan Ma and Qinsi Wang and Yuzhe Fu and Ming-Yu Chung and Yueqian Lin and Zhijian Liu and Jianyi Zhang and Danyang Zhuo and Yiran Chen},
      year={2025},
      eprint={2510.12872},
      archivePrefix={arXiv},
      primaryClass={cs.MA},
      url={https://arxiv.org/abs/2510.12872}, 
}

@article{molmo2024,
  title={Molmo and PixMo: Open Weights and Open Data for State-of-the-Art Vision-Language Models}, 
      author={Matt Deitke and Christopher Clark and Sangho Lee and Rohun Tripathi and Yue Yang and Jae Sung Park and Mohammadreza Salehi and Niklas Muennighoff and Kyle Lo and Luca Soldaini and Jiasen Lu and Taira Anderson and Erin Bransom and Kiana Ehsani and Huong Ngo and YenSung Chen and Ajay Patel and Mark Yatskar and Chris Callison-Burch and Andrew Head and Rose Hendrix and Favyen Bastani and Eli VanderBilt and Nathan Lambert and Yvonne Chou and Arnavi Chheda and Jenna Sparks and Sam Skjonsberg and Michael Schmitz and Aaron Sarnat and Byron Bischoff and Pete Walsh and Chris Newell and Piper Wolters and Tanmay Gupta and Kuo-Hao Zeng and Jon Borchardt and Dirk Groeneveld and Crystal Nam and Sophie Lebrecht and Caitlin Wittlif and Carissa Schoenick and Oscar Michel and Ranjay Krishna and Luca Weihs and Noah A. Smith and Hannaneh Hajishirzi and Ross Girshick and Ali Farhadi and Aniruddha Kembhavi},
      year={2024},
      eprint={2409.17146},
      archivePrefix={arXiv},
      primaryClass={cs.CV},
      url={https://arxiv.org/abs/2409.17146}, 
}

@misc{tong2024eyeswideshutexploring,
    title={Eyes Wide Shut? Exploring the Visual Shortcomings of Multimodal LLMs}, 
      author={Shengbang Tong and Zhuang Liu and Yuexiang Zhai and Yi Ma and Yann LeCun and Saining Xie},
      year={2024},
      eprint={2401.06209},
      archivePrefix={arXiv},
      primaryClass={cs.CV},
      url={https://arxiv.org/abs/2401.06209}, 
}

\appendix

\section{Results on Qwen2.5-VL-7B}
\label{app:7b}
To verify that our conclusions are not specific to the 3B backbone, we repeat the main evaluation on Qwen2.5-VL-7B with the same training recipe (Table~\ref{tab:7b}). Absolute accuracy rises across the board; the qualitative conclusions from the 3B experiments (consistent gains over long-CoT, self-consistency, Para-Thinker, and multi-agent-debate baselines, with the largest improvement on HallusionBench) carry over to 7B, and HallusionBench remains the benchmark on which \method most clearly separates from the strongest baseline.

\begin{table*}[t]
  \caption{Results on the Qwen2.5-VL-7B backbone with the same training recipe. The qualitative conclusions from the 3B experiments carry over to 7B. ``MAC'' denotes Multi-Agent Capability Injection (Stage~1) and ``RDMAO'' denotes Role-Decoupled Multi-Agent Optimization (Stage~2); \underline{underline} marks the Stage-1-only intermediate checkpoint and \textbf{bold} marks the final two-stage \method.}
  \label{tab:7b}
  \centering
  \small
  \setlength{\tabcolsep}{5pt}
  \begin{tabular}{l c c c c c c}
    \toprule
    \textbf{Method} & \textbf{V*} & \textbf{CountB.} & \textbf{RefCOCO} & \textbf{RefCOCO+/g} & \textbf{Hallus.} & \textbf{Tok.} \\
    \midrule
    Base MLLM (greedy, 7B)               & 64.5             & 56.4             & 79.7             & 73.1 / 74.6                         & 47.8             & 134  \\
    + Long CoT                           & 68.7             & 59.6             & 80.8             & 74.5 / 75.9                         & 50.6             & 605  \\
    + Self-Consistency ($K{=}4$)         & 70.4             & 62.0             & 81.7             & 75.1 / 76.4                         & 52.3             & 2410 \\
    + Para-Thinker ($K{=}4$)             & 72.5             & 63.5             & 82.6             & 76.0 / 77.2                         & 53.9             & 2456 \\
    + Multi-Agent Debate                 & 73.6             & 64.5             & 83.0             & 76.4 / 77.6                         & 55.4             & 2854 \\
    \midrule
    \method (MAC only, 7B)               & \underline{74.9} & \underline{66.7} & \underline{84.4} & \underline{77.6} / \underline{78.6} & \underline{58.0} & 2503 \\
    \textbf{\method (MAC+RDMAO, 7B)}     & \textbf{80.2}    & \textbf{71.5}    & \textbf{88.1}    & \textbf{82.0} / \textbf{82.9}       & \textbf{64.7}    & 2540 \\
    \bottomrule
  \end{tabular}
\end{table*}

\section{Training data and hyper-parameters}
\label{app:hparam}

\paragraph{Backbone, training stages, and protocol.}
We instantiate \method on a Qwen2.5-VL-3B~\citep{bai2025qwen25vltechnicalreport} backbone, with a complementary 7B run reported in Appendix~\ref{app:7b}. Training proceeds in two stages: Stage~1 performs Multi-Agent Capability Injection under the context-isolation mask of Eq.~\ref{eq:sft}; Stage~2 continues from the Stage-1 checkpoint and performs Role-Decoupled Multi-Agent Optimization with on-policy DAPO~\citep{yu2025dapo} using the role-specific advantages of Eq.~\ref{eq:adv}. All main experiments instantiate the framework with a fixed four-path collaboration protocol, chosen to align with natural visual partitions and complementary scan-order reasoning. The Worker advantage weight $\lambda$ is selected on validation as in Table~\ref{tab:lambda}. Train and validation splits are disjoint, with validation used exclusively for model selection and final evaluation.

\paragraph{Evaluation protocol details.}
Unless explicitly stated otherwise, all reported numbers are computed on the official validation or test splits provided by the corresponding benchmarks, using the benchmark-standard metrics and post-processing rules. For open-ended generation tasks, the final answer is extracted from the Summary Agent's output using the same normalization rule across methods. For grounding tasks, we follow the benchmark convention for box or region matching. Each timing measurement in the efficiency table is averaged over repeated runs on the same hardware. We interpret small absolute differences conservatively and do not claim significance beyond the resolution supported by these repeated measurements.

\paragraph{SFT data construction.}
Stage-1 traces are synthesised by a stronger MLLM teacher. The exact teacher system is withheld for double-blind submission and will be named in the camera-ready release together with the data-generation prompts. For each $(v,q)$ pair, the teacher is prompted to produce $K{+}2$ aligned segments: a Main Agent task allocation chosen by task type from Block-based and Scan-order patterns, $K$ Worker Agent traces each targeting a distinct visual sub-problem, and a Summary Agent trace that reconciles the Worker outputs into a final answer conditioned on the visual context. For the running V\textsuperscript{\textasteriskcentered}-style example, the Worker sub-problems correspond to layout, counting, spatial relation, and verification. The segments are stored as one packed multi-agent trajectory annotated with role tokens, which the Stage-1 context-isolation mask of Eq.~\ref{eq:sft} then enforces. We filter the synthesised pool with three conservative rules: (i) reject malformed role sequences, including missing or duplicated role tokens and fewer than $K$ Worker Agent traces; (ii) reject traces with no extractable final answer in the Summary segment; (iii) when a benchmark gold answer is available, reject traces whose Summary answer disagrees with the gold. We additionally stratify the retained pool by task family, covering visual search, counting, grounding, and hallucination, and by visual-token bin so that no single regime dominates. The total number of Stage-1 trajectories used in this paper is approximately $163{,}000$; training and validation splits are disjoint and validation is used exclusively for model selection.

\paragraph{RL hyper-parameters.}
We summarize the concrete training configuration used in the Stage-2 (Role-Decoupled Multi-Agent Optimization) script. The backbone is initialized from a Qwen2.5-VL-3B Stage-1 checkpoint. The RL trainer uses a fully on-policy DAPO variant in which $\texttt{train\_batch}{=}\texttt{ppo\_mini\_batch}{=}16$ and the number of PPO epochs per on-policy batch is $1$; the per-GPU micro-batch size is $1$ for both policy optimization and log-probability computation. The Worker advantage weight (the parameter named \texttt{dtpo\_lambda} in the training config; corresponds to $\lambda$ in Eq.~\ref{eq:adv}) is set to $0.5$ in this configuration. The optimizer learning rate is $5\times10^{-7}$. KL is applied as a loss with coefficient $0.01$ and low-variance KL type, while the reward-side KL penalty is disabled. Entropy coefficient is $0.02$, clipping thresholds are $0.20$ and $0.28$, the auxiliary clip parameter is $3.0$, and the advantage clip is $5.0$.

The data loader uses $4$ workers. The maximum prompt length is $18{,}000$ tokens and the maximum response length is $2{,}048$ tokens. Rollout uses vLLM with temperature $1.0$, $n=8$ generated samples per group, block size $16$, maximum batched tokens $64{,}000$, tensor-parallel size $1$, disabled chunked prefill, eager execution enabled, and GPU memory utilization target $0.4$. The native multi-agent inference engine is enabled with four role-trigger tokens and \texttt{parthink\_size}=4 (one trigger per Worker Agent). Dynamic group filtering is enabled using the outcome reward as the filtering metric, with up to $100$ generated batches considered during filtering; concretely, to keep the Summary Agent from degenerating into a naive majority voter we retain only Stage-2 samples with at most two correct Worker Agents out of $K{=}4$, and we stratify by task family and by visual-token bin for data balance. The training run uses $8$ GPUs on a single node and saves checkpoints every $50$ steps.

\paragraph{Hardware and software.}
The experiments are run on a single machine equipped with $8\times$ NVIDIA H20 GPUs. The rollout engine uses float16 generation, and the training system enables padding removal for the actor model, parameter offloading for the reference model, and a custom vLLM-based scheduler to support sequence fork/merge for the multi-agent collaboration protocol. These implementation choices are important for reproducing the inference time and throughput reported in Table~\ref{tab:efficiency}.

\section{Extended ablations}
\label{app:abl}
The sensitivity of the decoupling weight $\lambda$ is reported in Table~\ref{tab:lambda}.

\begin{table}[t]
  \caption{Sensitivity to the Worker advantage weight $\lambda$ in the role-decoupled advantage objective (Eq.~\ref{eq:adv}). $\lambda{=}0$ recovers an outcome-only RL run (only $A_\text{out}$ is propagated to Worker tokens); large $\lambda$ over-weights the Worker reward and amplifies its noise. We use $\lambda{=}0.5$ for all main-text experiments.}
  \label{tab:lambda}
  \centering
  \small
  \begin{tabular}{c c c c}
    \toprule
    $\lambda$ & \textbf{V*} & \textbf{CountBench} & \textbf{HallusionBench} \\
    \midrule
    $0.0$           & 69.7          & 62.8          & 54.6          \\
    $0.25$          & 71.9          & 65.1          & 57.3          \\
    $\mathbf{0.5}$  & \textbf{72.8} & \textbf{66.2} & \textbf{59.0} \\
    $1.0$           & 72.4          & 65.7          & 58.4          \\
    $2.0$           & 70.8          & 64.0          & 56.9          \\
    \bottomrule
  \end{tabular}
\end{table}

\section{Task Allocation Patterns: Details and Data Examples}
\label{app:patterns}

This appendix expands on the two fixed allocation patterns introduced in \S\ref{sec:mainagent} and shows two concrete Stage-1 (MAC) data examples used to teach the shared policy how each pattern is realised.

\subsection{Block-based Allocation}
\label{app:block}

\paragraph{Pattern definition.}
Given an image $v$ and the four-path collaboration protocol used in this paper, the Main Agent partitions $v$ into $K{=}4$ disjoint visual blocks. By default the blocks correspond to the four image quadrants (top-left, top-right, bottom-left, bottom-right). For tasks where the natural visual structure is not aligned with quadrants (e.g., elongated horizontal scenes or images with a strong centre bias), the Main Agent emits a task-specific partition---for instance, a $1{\times}4$ horizontal strip layout, a $2{\times}2$ centre-vs-periphery layout, or a foreground/background separation---chosen from a small fixed pool. Worker Agent $i$ is then conditioned on $v$, $q$, the Main Agent's segmentation $m$, and its role token $r_{\text{worker}_i}$, and is asked to collect local visual evidence within its assigned block while ignoring tokens from sibling Workers (Eq.~\ref{eq:sft}).

\paragraph{When this pattern fits.}
Block-based allocation is well suited to tasks dominated by local visual evidence collection: high-resolution visual search where the answer is contained in one region (e.g., V*), referring-expression grounding (RefCOCO/+/g), fine-grained perception (MMVP), and attribute or hallucination verification (HallusionBench). The disjointness of the blocks provides a hard inductive bias that complementary regions are scanned by complementary Workers, which the Stage-1 context-isolation mask then consolidates by preventing one Worker from copying another's reasoning.

\paragraph{Data example.}
Stage-1 training data follow the packed multi-agent trajectory format introduced in \S\ref{sec:formulation}. The teacher first generates per-block captions, the four blocks are then linked into separate Worker reasoning paths via $\langle\text{think}_i\rangle/\langle/\text{think}_i\rangle$ tags, and the Summary Agent's reconciled answer is wrapped in $\langle\text{summary}\rangle/\langle/\text{summary}\rangle$. We then apply the conservative filtering rules in Appendix~\ref{app:hparam} (malformed sequences, missing answers, gold-answer disagreement) before training. A representative MMVP-style block-based example is shown below.

\begin{sftbox}
  \begin{center}
    \includegraphics[width=0.35\linewidth,height=3cm,keepaspectratio]{figures/COCO_train2014_000000000009.jpg}
  \end{center}
  \texttt{<|User|>}\textit{<image> What is the color of the container holding the broccoli and meatball? A. yellow B. pink C. blue D. green. This is a multiple-choice question, please select one option. You FIRST think about the reasoning process as an internal monologue and then summarize the reasoning process to get the final answer. The summary process MUST BE enclosed within \texttt{<summary>\,</summary>} tags. The final answer MUST BE put in \texttt{\textbackslash boxed\{\}}.}\texttt{<|Assistant|><vllm\_pad>}

  \medskip
  \texttt{<think1>} \\
  Top-Left: This quadrant shows a pink container holding a slice of white bread with a dollop of white spread, several whole almonds, and two dried fruit slices, one of which is round and brown. The container has a red paper liner.\\
  \texttt{</think1>}

  \medskip
  \texttt{<think2>} \\
  Top-Right: This quadrant shows a pink container filled with yellow fruit pieces, including what appear to be pineapple chunks and mandarin orange segments. The container is set against a blue background.\\
  \texttt{</think2>}

  \medskip
  \texttt{<think3>} \\
  Bottom-Left: This quadrant shows a yellow container holding a round, brown meatball covered in red sauce and a large floret of green broccoli. The container has a yellow rim.\\
  \texttt{</think3>}

  \medskip
  \texttt{<think4>} \\
  Bottom-Right: This quadrant shows a dark brown, round cookie with a textured surface. Part of a blue container edge is visible, and another similar cookie is partially visible behind the first one.\\
  \texttt{</think4>}

  \medskip
  \texttt{<summary>} \\
  By analyzing the four parallel reasoning paths above, only Worker~3 reports a yellow container that simultaneously holds a meatball and broccoli. The final answer is \fbox{A. yellow}.\\
  \texttt{</summary>}
\end{sftbox}

\subsection{Scan-order Allocation}
\label{app:scan}

\paragraph{Pattern definition.}
Under Scan-order allocation, the Main Agent does not partition the image; instead it assigns each Worker Agent a distinct global scanning trajectory. The four trajectories used in this paper are left-to-right, right-to-left, top-to-bottom, and bottom-to-top. Each Worker Agent retains a global receptive field over the full image and traverses it in its assigned order, enumerating or verifying objects along the way. The disagreements between scan orders, rather than spatial coverage, become the source of complementary evidence.

\paragraph{When this pattern fits.}
Scan-order allocation is well suited to tasks where the dominant error mode is missing or double-counting an instance under a single scanning direction: precise object counting (Pixmo, CountBench), global verification (e.g., \emph{is every object in the image still present after a perturbation?}), and spatial-consistency checking (left-of, above-of relations). Permuting the traversal order across Workers is a cheap way to expose order-dependent miscounting at training time, which the Summary Agent then learns to reconcile rather than majority-vote on.

\paragraph{Data example.}
Stage-1 traces under Scan-order allocation use the same packed trajectory format and the same filtering rules as Block-based traces; the only difference is that each $\langle\text{think}_i\rangle$ block enumerates the same scene under a distinct scanning order. A representative counting example is shown below.

\begin{sftboxone}
  \begin{center}
    \includegraphics[width=0.35\linewidth,height=3cm,keepaspectratio]{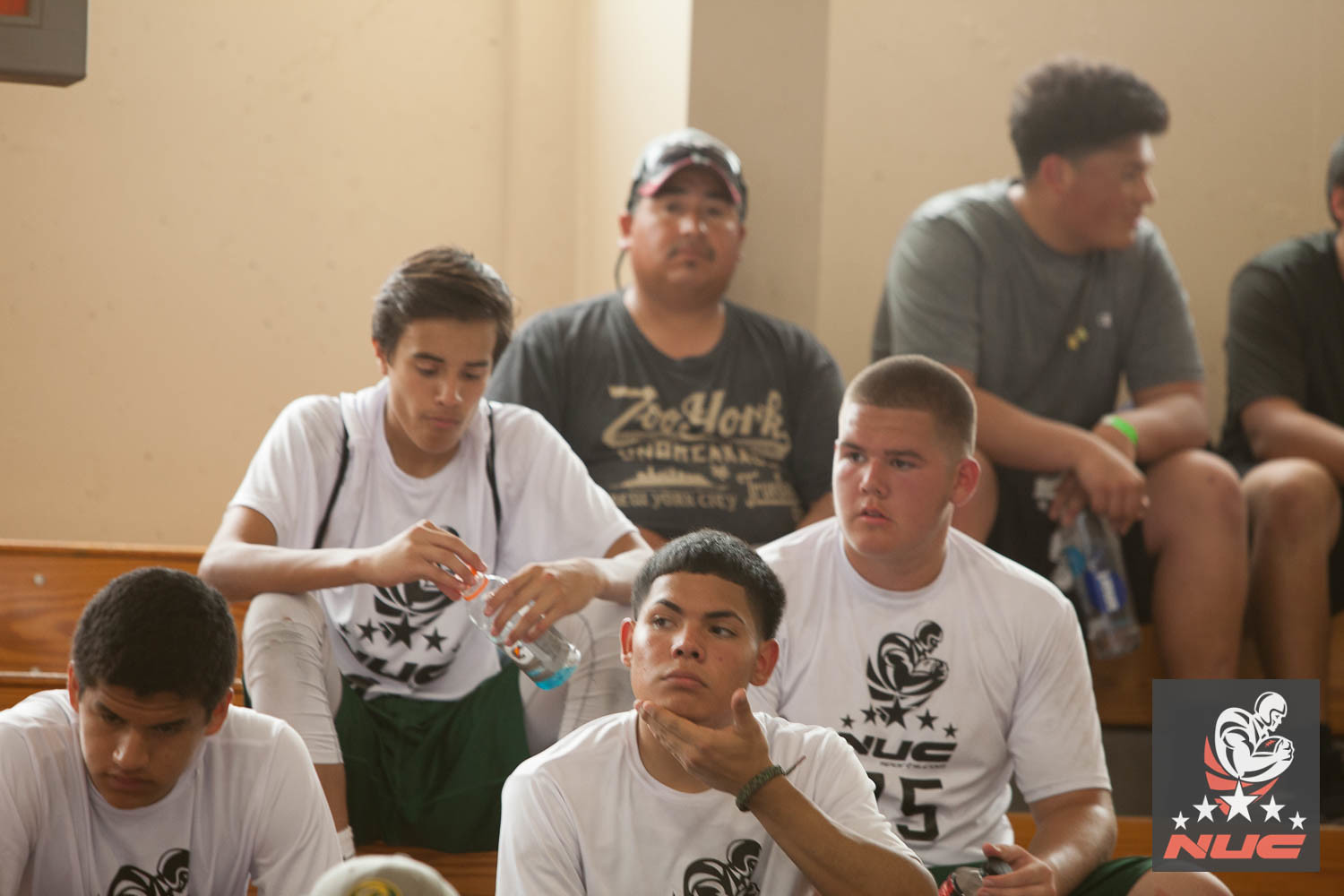}
  \end{center}
  \texttt{<|User|>}\textit{<image> How many people are in the image? You FIRST think about the reasoning process as an internal monologue. In your thoughts, try to locate and count the objects one by one or group by group to ensure accuracy. Then, summarize the counting process to get the final number. The summary process MUST BE enclosed within \texttt{<summary>\,</summary>} tags. The final answer MUST BE a single integer put in \texttt{\textbackslash boxed\{\}}.}\texttt{<|Assistant|><vllm\_pad>}

  \medskip
  \texttt{<think1>} \\
  Scanning Left-to-Right: 1.~Found a young man in a white shirt looking down on the far left. That makes 1.~2.~Next, a young man in a white shirt pouring water from a bottle. That makes 2.~3.~Behind him, a man in a dark grey ``Zoo York'' t-shirt and a cap. That makes 3.~4.~In the centre foreground, a young man in a white shirt with his hand on his chin. That makes 4.~5.~To his right, a young man in a white shirt with a ``NUC'' logo. That makes 5.~6.~Behind him, a young man in a grey t-shirt looking to the right. That makes 6.~7.~On the far right, a partially visible person in a dark shirt. That makes 7. Total count: 7.\\
  \texttt{</think1>}

  \medskip
  \texttt{<think2>} \\
  Scanning Top-to-Bottom: Starting from the top, I first encounter the man in the ``Zoo York'' t-shirt and cap. That makes 1. To his right, the young man in the grey t-shirt. That makes 2. To the far right, the partially visible person in a dark shirt. That makes 3 \dots~Total count: 7.\\
  \texttt{</think2>}

  \medskip
  \texttt{<think3>} \\
  Scanning Right-to-Left: Starting from the far right, I see a partially visible person in a dark shirt. That makes 1. Moving left, I see a young man in a grey t-shirt. That makes 2. Next, a young man in a white ``NUC'' shirt. That makes 3. In front of him, a young man in a white shirt with his hand on his chin. That makes 4. Behind him, a man in a ``Zoo York'' t-shirt and cap. That makes 5 \dots~Total count: 7.\\
  \texttt{</think3>}

  \medskip
  \texttt{<think4>} \\
  Scanning Bottom-to-Top: Starting from the bottom, I first find the young man looking down on the far left. That makes 1. Moving up, I see the young man with his hand on his chin. That makes 2. To his right, the young man in the ``NUC'' shirt. That makes 3. Above him, the young man in the grey t-shirt \dots~Total count: 7.\\
  \texttt{</think4>}

  \medskip
  \texttt{<summary>} \\
  All four scans converge on the same count: every individual is enumerated exactly once across the four orderings, which mutually verify against double-counting and omission. The final answer is \fbox{7}.\\
  \texttt{</summary>}
\end{sftboxone}

\subsection{Pattern selection at inference time}
\label{app:select}

By default the Main Agent emits Block-based allocation. It switches to Scan-order allocation when the task is dominated by counting or by global verification. The selection signal is a short tag inside the Main segment $m$, learned during Stage-1 from the teacher-annotated allocation choice, so all four Worker Agents and the Summary Agent see the chosen pattern as part of their context. We deliberately keep the pattern set small (two patterns, four canonical scan orders, four fixed quadrant blocks) so that the role-decoupled optimisation in \S\ref{sec:rdmao} operates on a small number of stable inductive biases rather than on a free-form planner output.

\end{document}